\documentclass{article}

\usepackage[nonatbib,preprint]{neurips_2025}

\usepackage[numbers]{natbib}
\usepackage[utf8]{inputenc} 
\usepackage[T1]{fontenc}    
\usepackage{hyperref}       
\usepackage{url}            
\usepackage{booktabs}       
\usepackage{amsfonts}       
\usepackage{nicefrac}       
\usepackage{microtype}      
\usepackage{xcolor}         
\usepackage[inline]{enumitem}
\usepackage{graphicx}    
\usepackage{makecell}    
\usepackage{multirow}
\usepackage{overpic}
\usepackage{float}
\usepackage{wrapfig}
\usepackage{xcolor,colortbl}
\usepackage{amsmath}
\usepackage{amssymb}
\usepackage{subcaption}
\usepackage{caption}
\usepackage{pifont}
\usepackage{xspace}
\usepackage{ragged2e}
\usepackage[skins,breakable]{tcolorbox}

\usepackage{lipsum}

\newcommand{\graytext}[1]{\noindent \textcolor[rgb]{0.6, 0.6, 0.6}{{#1}}}

\newcommand{\designbenchmark}{\textsc{Design-Multi-Layer-Bench}\xspace}
\newlength\savewidth\newcommand\shline{\noalign{\global\savewidth\arrayrulewidth
\global\arrayrulewidth 1pt}\hline\noalign{\global\arrayrulewidth\savewidth}}

\newcommand{\hl}{\cellcolor{gray!10}}

\newcommand{\tablestyle}[2]{\setlength{\tabcolsep}{#1}\renewcommand{\arraystretch}{#2}\centering\footnotesize}

\makeatletter
\renewcommand{\paragraph}{%
  \@startsection{paragraph}{4}{\z@}%
                {0.0ex \@plus 0.3ex \@minus 0.1ex}%
                {-1em}%
                {\normalsize\bf}%
}
\makeatother

\usepackage{tikz}
\usepackage{xcolor}

\title{PixelART: Image-to-Layer Decomposition without Latents or Text-to-Image Pretraining}
\author{%
  \textbf{Zelin Jia$^{1}$\quad Zhao Zhang$^{2}$\quad Zhicong Tang$^{2}$ \quad Yuhui Yuan$^{2*}$\quad Shixia Liu$^{1}$}\thanks{Corresponding author.}\vspace{3pt} \\
  $^1$School of Software, BNRist, Tsinghua University ~~\quad\quad $^2$Canva Research
\vspace{3pt} \\
  \texttt{\small ryanyuan@canva.com, shixia@tsinghua.edu.cn}\\
}

\begin{document}

\maketitle

\begin{figure}[H]
\centering
\vspace{-5mm}
\includegraphics[width=0.98\linewidth]{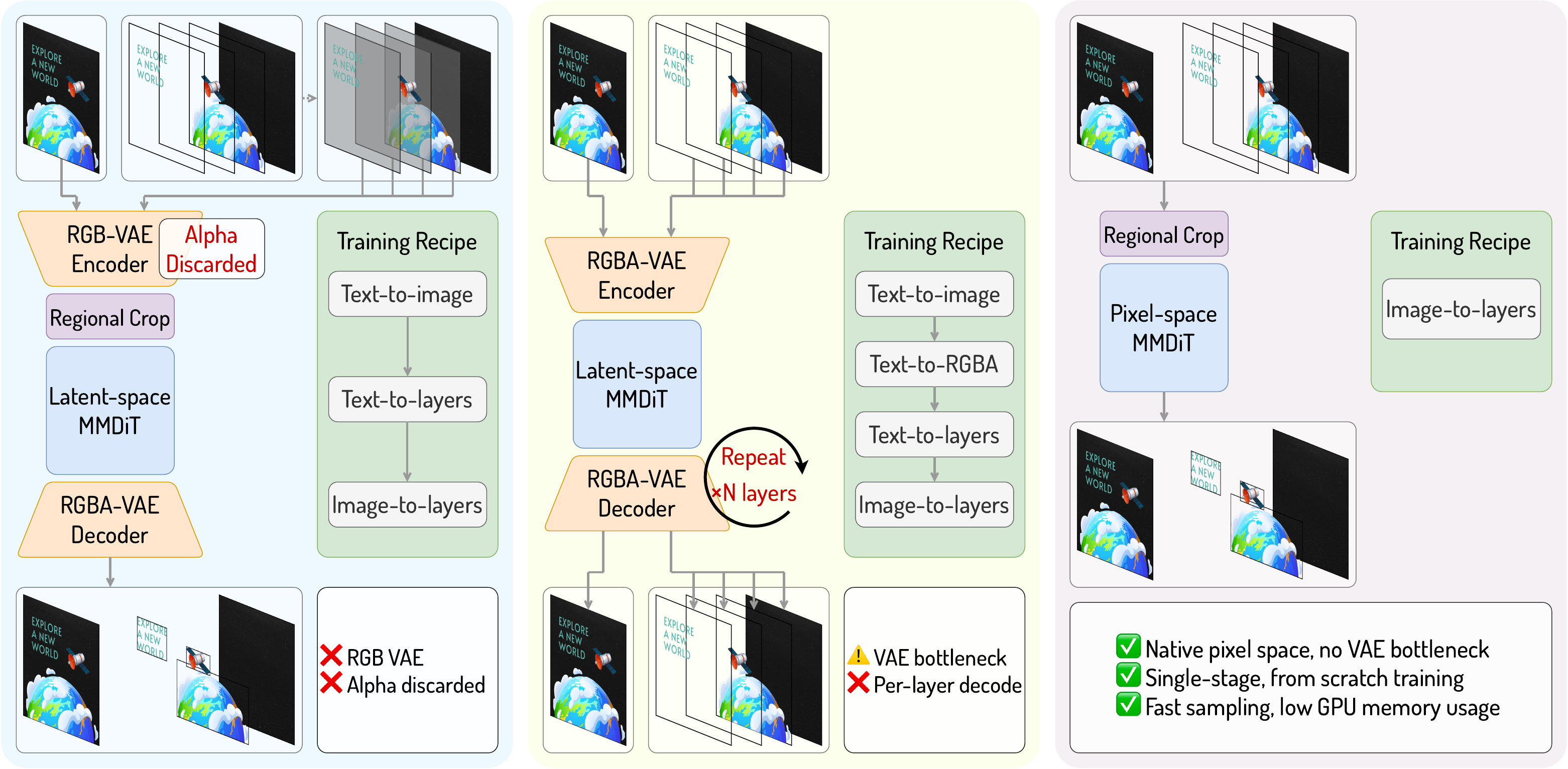}
\vspace{-1mm}
\caption{\footnotesize Comparison with previous Image-to-Layer decomposition systems. Left: ART~\cite{pu2025art} and its variant; Middle: Qwen-Image-Layered~\cite{yin2025qwenimagelayered}; Right: Ours. Our PixelART is much simpler and trained from scratch.}
\label{fig:teaser}
\vspace{-4mm}
\end{figure}

\begin{abstract}

Image-to-layer decomposition converts a flattened image into editable RGBA layers, enabling element-level editing in design workflows. 
Existing diffusion-based systems typically adapt large pretrained text-to-image (T2I) models and introduce RGBA autoencoders or variable-layer architectural modules. 
We revisit this design choice and ask whether layer decomposition truly requires these heavyweight components. 
We introduce PixelART, a pixel-space rectified-flow Transformer trained from scratch for image-to-layer (I2L) decomposition. PixelART directly denoises regional RGBA pixel patches using a single-stream multi-modal diffusion Transformer, avoiding RGBA-VAEs, pretrained T2I backbones, and layer-specific decoders. 
We identify a key property of the task: high-noise timesteps determine layer assignment and coarse layer organization, while low-noise timesteps mainly refine color, alpha, texture, and boundaries. 
Based on this observation, we propose a \emph{terminal-boosted timestep sampling} strategy to increase training coverage in the high-noise layer assignment regime. 
Trained on 4M multi-layer design templates, PixelART achieves state-of-the-art layer decomposition and composite reconstruction on \designbenchmark\space and LICA with over $80\%$ fewer parameters, $98\%$ lower latency, and $85\%$ lower memory than the recent Qwen-Image-Layered model. 
Ablation experiments show that pixel-space $\mathbf{x}$-prediction, terminal-boosted timestep sampling, and data/model scaling are critical, while T2I pretraining brings marginal benefits to the I2L task.

\end{abstract}

\section{Introduction}

\begin{figure}[t]
    \centering
    \includegraphics[width=0.97\linewidth, trim=0cm 0cm 0cm 0cm]{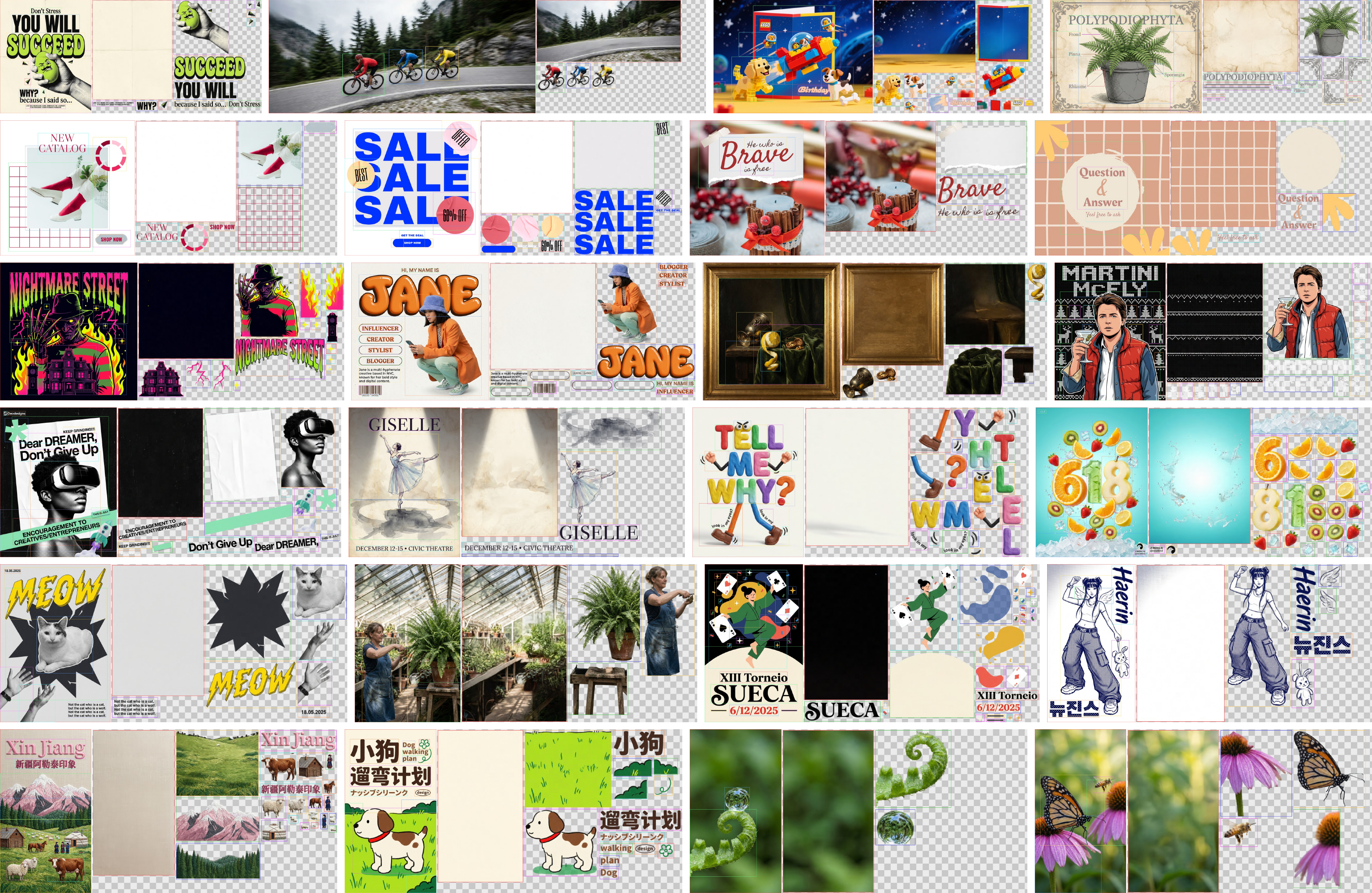}
    \caption{\small Image-to-layer decomposition results with our PixelART-H/32.}
    \label{fig:main_results}
    \vspace{-5mm}
\end{figure}

Flat image generators, such as Nano-Banana~\cite{nanobanana,nanobanana2,nanobananapro} and GPT-Image~\cite{openai_gpt_image_1,openai_gpt_image_1_5,openai_gpt_image_2}, can synthesize visually rich raster images, but their outputs remain difficult to edit. 
In contrast, professional design tools rely on layered representations, where text, graphics, backgrounds, shadows, and decorative elements can be edited independently~\cite{li2026chartgalaxy}. 
Image-to-layer (I2L) decomposition aims to recover this editable structure by converting a flattened RGB image into an ordered stack of RGBA layers, enabling element-level editing, restyling, and recomposition.
Recent diffusion-based systems, including ART~\cite{pu2025art}, Qwen-Image-Layered~\cite{yin2025qwenimagelayered}, OmniPSD~\cite{liu2025omnipsd}, and OmniAlpha~\cite{yu2025omnialpha}, have made rapid progress on layered generation and decomposition. 
Despite their differences, they largely follow a common recipe: adapt a large pretrained text-to-image (T2I) backbone, extend an RGB autoencoder into an RGBA autoencoder, and add task-specific mechanisms for transparent and variable-layer outputs.
While natural, this design is complex, and the latent autoencoder can become a bottleneck when modeling design layers with transparency, small text, sparse strokes, flat colors, and sharp alpha boundaries.\looseness=-1

We revisit this recipe and ask \emph{whether} strong I2L performance can be obtained without latent autoencoding or T2I \emph{pretraining}.
To this end, we propose PixelART, a pixel-space rectified-flow Transformer trained from scratch. 
PixelART directly denoises regional RGBA pixel patches using a single-stream Transformer, avoiding RGBA-VAEs, pretrained T2I backbones, layer-specific decoders, and variable-layer modules (Figure~\ref{fig:teaser}). 
We formulate the I2L task as a pixel-native assignment-and-completion problem, in which the model must assign visible pixels to layers, complete occluded content, and refine RGBA values.
We observe that layer assignment and detail refinement occur at different stages of the diffusion trajectory. 
High-noise timesteps determine layer assignment and structure; low-noise timesteps refine appearance like color, alpha, texture, and boundaries.
We propose a novel \emph{terminal-boosted timestep sampling} strategy to increase training coverage in the high-noise assignment regime.

PixelART achieves state-of-the-art layer decomposition and composite reconstruction quality on \designbenchmark and LICA while using substantially fewer parameters, lower latency, and memory than VAE-based diffusion baselines.
Ablations show that pixel-space $\mathbf{x}$-prediction, high-noise timestep coverage, and data/model scaling are all critical, whereas T2I initialization reduces early loss but yields marginal gain under our data-rich I2L setup.
Figure~\ref{fig:main_results} presents additional I2L decomposition visual results.
The main contributions of our work are:
\begin{itemize}[left=3mm, itemsep=0pt, topsep=0pt, parsep=2pt, partopsep=0pt]
    \item We show that I2L decomposition can be effectively solved by a pixel-space diffusion Transformer trained from scratch, without latent autoencoding or T2I pretraining.
    \item We introduce PixelART, a single-stream regional Transformer that directly predicts editable RGBA pixel patches without layer-specific decoders or variable-layer modules.
    \item We identify the importance of high-noise timesteps for layer assignment and propose terminal-boosted timestep sampling to improve decomposition reliability. System-level comparisons and ablations demonstrate that PixelART matches or outperforms larger T2I-initialized pipelines while maintaining a simpler and more efficient design.
\end{itemize}
\section{Related Work}

\paragraph{Pixel-space Diffusion Models.}
Pixel-space diffusion has evolved through three waves.
The \emph{first wave} operates directly in RGB pixel space with U-Net backbones, including DDPM~\cite{ho2020ddpm}, ADM~\cite{dhariwal2021adm}, GLIDE~\cite{nichol2022glide}, Imagen~\cite{saharia2022imagen}, DALL-E~2~\cite{ramesh2022dalle2}, and Cascaded Diffusion~\cite{ho2022cascaded}, all relying on overcomplete channels and dense skip connections to cope with high dimensionality.
The \emph{second wave} shifted denoising into a compressed VAE latent, including LDM~\cite{rombach2022ldm}, SDXL~\cite{podell2023sdxl}, eDiff-I~\cite{balaji2022ediffi}, DALL-E~3~\cite{betker2023dalle3}, FLUX~\cite{flux2024}, SD3~\cite{esser2024sd3}, and Qwen-Image~\cite{wu2025qwenimage}, dramatically lowering per-step compute but tying generation quality to VAE capacity—a poor fit for downstream tasks like layer decomposition whose inputs and targets are pixel-native.
The \emph{third wave} revisits pixel space with Transformer backbones. SiD2~\cite{sid2}, PixelFlow~\cite{pixelflow}, and PixNerd~\cite{pixnerd} demonstrate that ViT-based pixel diffusion is competitive at high resolution when combined with loss weighting, cascaded multi-resolution training, and an auxiliary patch decoder, respectively. JiT~\cite{li2025jit}, pMF~\cite{lu2026onestep}, and PixelGen~\cite{pixelgen} go further, showing that a plain ViT with $\mathbf{x}$-prediction on large pixel patches matches latent diffusion without any tokenizer.
PixelART builds directly on this third wave: we inherit JiT's $\mathbf{x}$-prediction ViT and extend pixel-space diffusion from T2I to I2L decomposition, where the pixel-native input/output structure is an especially natural fit.

\paragraph{Multi-Layer Generation and Decomposition.}
Layered RGBA generation has mainly followed two paradigms. \emph{Simultaneous} methods generate all layers in a single forward pass, including Text2Layer~\cite{text2layer}, LayerDiff~\cite{layerdiff}, ART~\cite{pu2025art}, PrismLayer~\cite{prismlayer}, DreamLayer~\cite{huang2025dreamlayer}, and Qwen-Image-Layered~\cite{yin2025qwenimagelayered}. \emph{Sequential} methods generate or extract layers one at a time, including LayerDiffuse~\cite{zhang2024layerdiffuse}, COLE~\cite{cole}, OpenCOLE~\cite{opencole}, LayerD~\cite{suzuki2025layerd}, and OmniPSD~\cite{liu2025omnipsd}. Recent image-to-layer systems, such as Qwen-Image-Layered  and CLD~\cite{cld}, adapt pretrained text-to-image backbones with substantial task-specific engineering, including RGBA VAEs, variable-layer tokenization, region-conditioned Transformers, and multi-stage training. Commercial systems such as Magic Layers~\cite{magiclayers}, Lovart~\cite{lovart}, and Ideogram~\cite{ideogram} appear to follow similar recipes at product scale.
Adjacent work addresses matting, segmentation, and alpha-aware representation learning. Matting- and segmentation-based pipelines, including LayerD and trimap-free methods based on SAM~\cite{sam}, SAM2~\cite{sam2}, HQ-SAM~\cite{hqsam}, and ZIM~\cite{zim}, decompose images through cascaded detection, segmentation, matting, and inpainting, but lack end-to-end optimization and often struggle with hard boundaries or semi-transparent content. Transparency-focused VAE methods, including LayerDiffuse~\cite{zhang2024layerdiffuse}, AlphaVAE~\cite{wang2025alphavae}, Wan-Alpha~\cite{dong2025wanalpha}, and OmniAlpha~\cite{yu2025omnialpha}, extend RGB autoencoders to RGBA latent spaces for downstream generative models. Overall, prior approaches rely on T2I pretraining, RGBA-specific tokenizers, or cascaded expert modules. PixelART instead uses a single pixel-space diffusion Transformer trained from scratch, without a VAE or pretrained backbone, achieving competitive performance.

\section{Our Approach}
\label{sec:method}

\subsection{Image-to-Layer as Pixel-Space Inverse Generation}
\label{sec:i2l_formulation}

Most previous layered diffusion systems~\cite{pu2025art,cld,yin2025qwenimagelayered}
adapt pretrained text-to-image backbones, typically by extending an RGB VAE into an RGBA VAE and modifying an MMDiT backbone to support alpha transparency, variable layer numbers, and multiple resolutions.
In contrast, we formulate image-to-layer decomposition as a \emph{pixel-native inverse problem}: assigning visible pixels to editable layers and completing occluded content behind them.
These two subproblems are naturally defined in RGB/RGBA pixel space and do not inherently require text-to-image priors.

\paragraph{Formulation.}
Given a flattened RGB image
\(\mathbf I\in[0,1]^{H\times W\times 3}\),\footnote{All RGB and RGBA tensors are represented as normalized floating-point values in \([0,1]\) if not specified.} image-to-layer decomposition aims to recover an ordered stack of editable RGBA layers
\(\mathcal L=\{\mathbf L_0,\mathbf L_1,\ldots,\mathbf L_K\}\), where
\(\mathbf L_i\in[0,1]^{H\times W\times 4}\).
Here \(\mathbf L_0\) denotes the background layer and
\(\{\mathbf L_i\}_{i=1}^{K}\) denote the foreground layers in design z-order.
The predicted layers should reconstruct the input under alpha compositing:
\begin{equation}
\Pi_{\mathrm{RGB}}
\left(
\operatorname{Over}(\mathbf L_0,\mathbf L_1,\ldots,\mathbf L_K)
\right)
\approx \mathbf I ,
\end{equation}
where \(\operatorname{Over}(\cdot)\) denotes source-over compositing all transparent layers and the background layer in normalized RGBA space and
\(\Pi_{\mathrm{RGB}}\) extracts the RGB channels.

\paragraph{Regional layer parameterization.}
Following the anonymous regional formulation of ART~\cite{pu2025art}, we do not assign semantic labels to layer slots.
Each layer is specified only by a spatial region and its z-order.
Let
\(\mathcal R=\{r_0,r_1,\ldots,r_K\}\) denote the region set, where
\(r_0\) is the full canvas and \(r_i=(x_i,y_i,h_i,w_i)\) for \(i>0\) is the padded bounding box of the non-transparent support of layer \(i\).
During training, \(\mathcal R\) is obtained from the ground-truth design file.
At inference, \(\mathcal R\) can be provided by a user or produced by an automatic layout detector; PixelART is agnostic to the source of the regions.

For each RGBA crop
\(\mathbf X_i=\operatorname{crop}(\mathbf L_i,r_i)\in[0,1]^{h_i\times w_i\times4}\),
we pad it to a multiple of patch size \(P\) and form exact non-overlapping pixel patches:
\begin{equation}
\phi_P(\mathbf X_i)\in\mathbb R^{N_i\times 4P^2},
\qquad
N_i=
\left\lceil h_i/P \right\rceil
\left\lceil w_i/P \right\rceil .
\end{equation}
The clean target sequence is always defined in exact RGBA pixel-patch space:
\begin{equation}
\mathbf Y =
[
\phi_P(\mathbf X_0);
\phi_P(\mathbf X_1);
\ldots;
\phi_P(\mathbf X_K)
].
\end{equation}
The flattened input image is converted to normalized RGBA by appending an opaque alpha channel,
\(\widetilde{\mathbf I}=[\mathbf I;\mathbf 1]\).
We pad variable-length regional sequences and mask padding tokens in attention and loss.

\subsection{PixelART Architecture}
\label{sec:pixelart_architecture}

\paragraph{Pixel patch embedding.}
PixelART keeps the denoising target in exact pixel space but embeds patches into the Transformer width \(d\) for computation.
Let \(E(\cdot)\) denote the selected patch embedding.
We support two choices.
The default is a shared linear projection from \(4P^2\) to \(d\).
The second is a lightweight convolutional bottleneck, implemented as a \(P\times P\), stride-\(P\) convolution from \(4\) channels to \(d_b\), followed by a \(1\times1\) convolution from \(d_b\) to \(d\), following the spirit of 
JiT~\cite{li2025jit}.
This module only embeds pixel patches into the Transformer; it is not a VAE or latent decoder.
The output head still predicts exact RGBA pixel patches in \(\mathbb R^{N_i\times4P^2}\).

\paragraph{Anonymous regional tokens with 3D RoPE.}
PixelART treats all visual regions as anonymous RGBA patch sequences.
We do not provide regional prompts or use learned role embeddings to distinguish the input image, the background layer, or foreground layers, nor do we use layer-specific decoders.
For patch \(j\) from region \(r_i\), the initial Transformer token is $\mathbf h_{i,j}=E(\mathbf X_i)_j$.
Spatial and layer-slot information is injected only through 3D RoPE inside the attention layers.
Each visual token is assigned a 3D coordinate that specifies the $x$, $y$, and $z$ positions, where $x$ and $y$ correspond to the spatial position within the image, and $z$ refers to layer index.
For clean conditioning-image tokens, we use the full-canvas spatial coordinates and a single fixed layer slot reserved for the input image; for target layer tokens, $z_i$ follows the anonymous layer order.

\paragraph{Single-stream pixel diffusion Transformer.}
Inspired by single-stream diffusion Transformers~\cite{zimage2025}, PixelART concatenates optional text tokens, clean conditioning-image tokens, and noisy target-layer tokens into one sequence:
\begin{equation}
\mathbf S_t =
[
\mathbf T;\,
\mathbf G;\,
E(\mathbf Y_t^{(0)});\,
E(\mathbf Y_t^{(1)});\,
\ldots;\,
E(\mathbf Y_t^{(K)})
],
\end{equation}
where \(\mathbf T\) denotes text tokens,
\(\mathbf G=E(\widetilde{\mathbf I})\) denotes clean conditioning-image tokens,
and \(\mathbf Y_t^{(i)}\) denotes the noised pixel-patch sequence of target region \(i\).
After lightweight modality-specific processors, we apply all multi-modal self-attention layers over the entire sequence, allowing every target layer to attend to the input image, text, and all other layers.
The output head is read only on target-layer positions:
\begin{equation}
\widehat{\mathbf Y}_{\theta}
=
f_{\theta}(\mathbf S_t,t,\mathcal R)_{\mathrm{target}} .
\end{equation}
Conditioning and target tokens are separated by the diffusion mask: conditioning tokens remain clean, while target layer tokens are noised and supervised.

\subsection{Learning Objective and Sampling}
\label{sec:learning_objective}

PixelART follows rectified-flow training with the convention that \(t=0\) denotes pure noise and \(t=1\) denotes clean data.
Given clean target pixel patches \(\mathbf Y\), Gaussian noise
\(\boldsymbol\epsilon\sim\mathcal N(\mathbf 0,\mathbf I)\), and timestep \(t\sim q(t)\), we define
$
\mathbf Y_t
=
t\mathbf Y+(1-t)\boldsymbol\epsilon$.
The model predicts clean pixel patches \(\widehat{\mathbf Y}_{\theta}\).
Following the $\mathbf{x}$-prediction with $\mathbf{v}$-loss formulation, we convert the clean prediction into a velocity estimate:
\begin{equation}
\widehat{\mathbf v}_{\theta}
=
\frac{\widehat{\mathbf Y}_{\theta}-\mathbf Y_t}{1-t},
\qquad
\mathbf v=\mathbf Y-\boldsymbol\epsilon.
\label{eq:velocity}
\end{equation}
We optimize the velocity-space flow-matching objective:
\begin{equation}
\mathcal L_{\mathrm{flow}}
=
\mathbb E_{\mathbf Y,\boldsymbol\epsilon,t}
\left[
\frac{1}{|\Omega|}
\sum_{j\in\Omega}
\left\|
\widehat{\mathbf v}_{\theta,j}-\mathbf v_j
\right\|_2^2
\right],
\end{equation}
where \(\Omega\) denotes valid non-padding target tokens.
Equivalently, this objective corresponds to clean-pixel prediction with a timestep-dependent weight \(1/(1-t)^2\).
We also incorporate an LPIPS perceptual loss based on both VGG and ConvNeXt-V2, following~\cite{lu2026onestep} for improved performance.

\begin{figure}[t]
    \centering
    \includegraphics[width=0.9\linewidth]{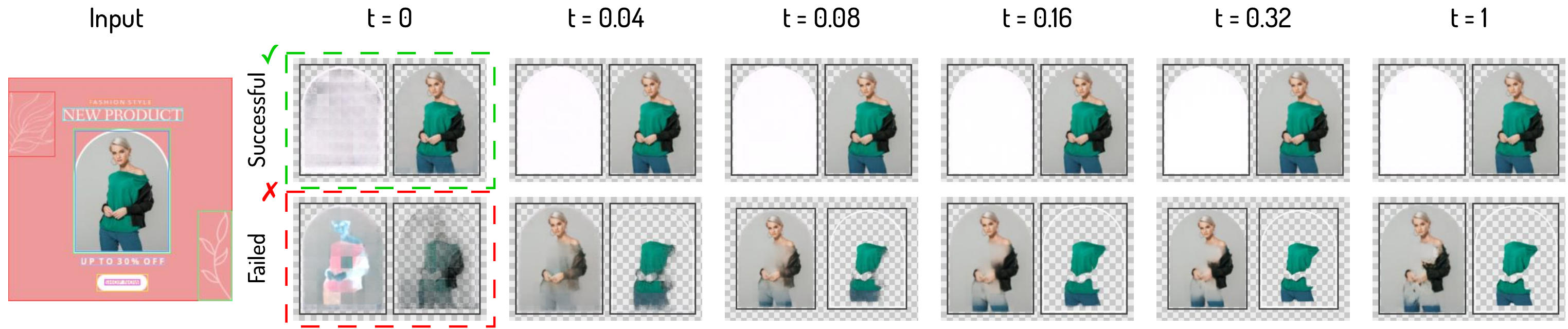}
\caption{\small 
\textbf{High-noise timesteps are critical for layer assignment.}
        We compare layer states along the sampling trajectory for checkpoints with and without terminal boosting.
        The boosted model (1st row) assigns regions correctly at the earliest high-noise stage, whereas the unboosted model (2nd row) makes a persistent layer-assignment error.
        Later steps mainly refine RGBA details.
        Green/red boxes mark correct/incorrect assignments.
}
\label{fig:timestep_illust}
\vspace{-3mm}
\end{figure}

\paragraph{Sampling and reconstruction.}
At inference, we feed the conditioning-image and text tokens, initialize the target pixel-patch sequence
\(\mathbf Y_0\sim\mathcal N(\mathbf 0,\mathbf I)\), and integrate the rectified-flow ODE from \(t=0\) to \(t=1\).
At each step \(t<1\), the model predicts clean patches \(\widehat{\mathbf Y}_{\theta}\) and estimates the velocity following Equation~\eqref{eq:velocity}.
We adopt classifier-free guidance (CFG)~\cite{ho2021classifier} in velocity space.
After sampling, each predicted region sequence \(\widehat{\mathbf Y}_i\) is unpatchified, cropped to remove padding, and pasted back to the canvas:
\begin{equation}
\widehat{\mathbf L}_i(\mathbf p)
=
\begin{cases}
\operatorname{clip}_{[0,1]}\!\left(\widehat{\mathbf X}_i(\mathbf p-\mathbf o_i)\right),
& \mathbf p\in r_i,\\
(0,0,0,0),
& \mathrm{otherwise},
\end{cases}
\end{equation}
where
\(\widehat{\mathbf X}_i=\operatorname{crop}_{h_i,w_i}(\phi_P^{-1}(\widehat{\mathbf Y}_i))\)
and \(\mathbf o_i=(x_i,y_i)\) is the origin of region \(r_i\).
The final output is the ordered editable RGBA stack
\(\{\widehat{\mathbf L}_0,\widehat{\mathbf L}_1,\ldots,\widehat{\mathbf L}_K\}\).

\subsection{Terminal-Boosted Timestep Sampling}
\label{sec:terminal_boost}

\paragraph{Logit-normal base schedule.}
Modern rectified-flow Transformers such as SD3, FLUX, and JiT~\cite{li2025jit} commonly sample timesteps from a logit-normal distribution~\cite{esser2024sd3,flux2024,wu2025qwenimage}.
A latent variable \(u\sim\mathcal N(\mu,\sigma^2)\) is mapped to \(t\in(0,1)\) by
\(t=\operatorname{sigmoid}(u)\), yielding density
\begin{equation}
q_{\mathrm{base}}(t)
=
\frac{1}{\sigma\sqrt{2\pi}}
\frac{1}{t(1-t)}
\exp\left(
-\frac{(\operatorname{logit}(t)-\mu)^2}{2\sigma^2}
\right).
\end{equation}
With \(t=0\) as pure noise and \(t=1\) as clean data, \(\mu\) controls which endpoint receives more mass; the standard logit-normal schedule concentrates mass at intermediate timesteps, with little near \(t=0\).

\paragraph{High-noise layer assignment.}
Image-to-layer decomposition differs from standard image generation because it must first solve a discrete assignment problem: which visible pixels belong to which output layer.
We interpret the output space as a disjoint union of manifolds, one for each valid layer assignment.
The model must choose the correct branch at the beginning of the sampling trajectory, when \(t\approx0\).
Once denoising has progressed, later steps mainly refine continuous RGBA values and often cannot switch to a different layer-assignment branch.
This motivates increasing the training probability of the high-noise region near \(t=0\). Figure~\ref{fig:timestep_illust} visualizes several examples illustrating failures caused by incorrect layer assignments at high-noise timesteps.

\begin{figure}[t]
\centering
\begin{minipage}[t]{0.66\linewidth}
\vspace{0pt}
\centering
\includegraphics[width=\linewidth]{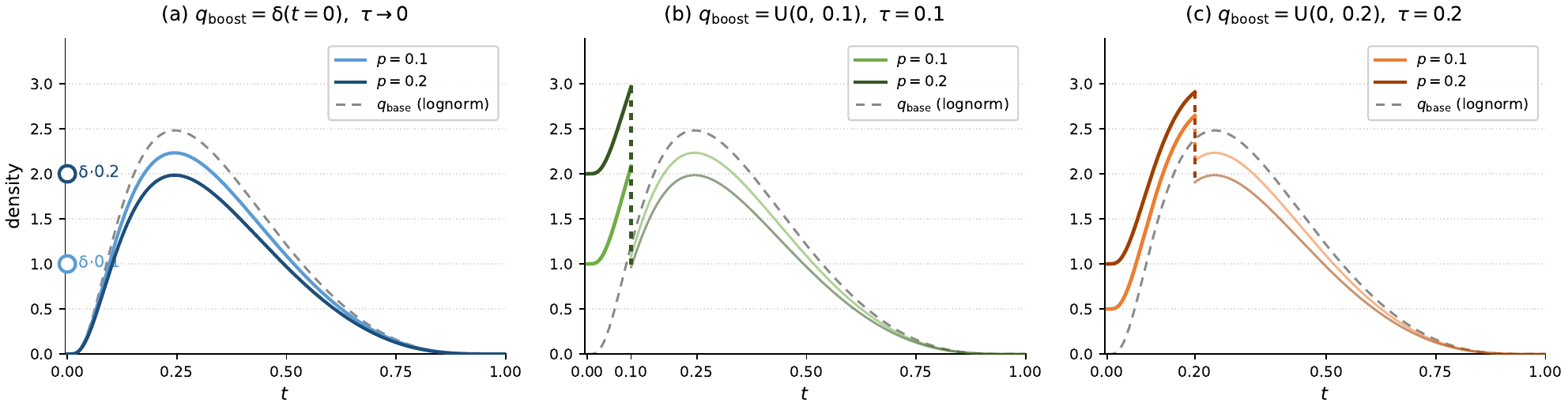}
\end{minipage}
\hfill
\begin{minipage}[t]{0.33\linewidth}
\vspace{0pt}
\caption{\small
\textbf{Terminal-boosted timestep distributions.}
We mix the base schedule with either a point mass at \(t=0\) or a uniform distribution \(\mathcal U(0,\tau)\) to increase the mass in the high-noise region.
}
\label{fig:timestep_density}
\end{minipage}
\vspace{-5mm}
\end{figure}

\paragraph{Terminal-boost formulation.}
We augment base schedule with an auxiliary high-noise component:
\begin{equation}
q(t)
=
p\cdot q_{\mathrm{boost}}(t;\tau)
+
(1-p)\cdot q_{\mathrm{base}}(t),
\label{eq:terminal_boost}
\end{equation}
where \(p\) is the boost probability and \(q_{\mathrm{boost}}\) is supported on \([0,\tau]\).
We consider two simple choices:
\begin{equation}
q_{\mathrm{boost}}(t;\tau)
=
\begin{cases}
\delta(t=0), & \tau\rightarrow0,\\
\mathcal U(0,\tau), & \tau>0.
\end{cases}
\end{equation}
The point-mass boost emphasizes the pure-noise endpoint, while the uniform boost spreads probability over an early high-noise interval.
Figure~\ref{fig:timestep_density} visualizes the resulting mixture density for
$p \in \{0.1, 0.2\}$ and $\tau \in \{0, 0.1, 0.2\}$.
All these variants allocate more probability mass to high-noise regions.
The compatibility of the terminal-boosted schedule with $\mathbf{x}$-prediction $\mathbf{v}$-loss in Supplementary~\ref{app:pred_space}.

\section{Experiments}
\label{sec:experiments}

\paragraph{Architecture.}
We instantiate three model variants, PixelART-B/L/H, following ViT~\cite{dosovitskiy2021vit} naming. Each variant has $12/24/32$ transformer blocks, $768/1024/1280$ hidden dimensions, $12/16/16$ attention heads, $128/256/256$ bottleneck dimensions, and $168\mathrm{M}/523\mathrm{M}/1.05\mathrm{B}$ parameters, respectively.

\vspace{-1.5mm}
\paragraph{Training setting.}
We conduct all ablation experiments with PixelART-B/16 (i.e., patch size p=16) at $256\times256$ resolution by default, and train each model for $100$K steps, \textit{i.e.} $20$ epochs and one day of training with $8\times$~H200 GPUs by default.
For system-level experiments, we train PixelART-H/32 with $32\times$ H200 GPUs first at $256\times256$ resolution with patch size $32$ for $80$ epochs, then progressively scale to $512\times512$ and $1024\times1024$ resolutions for additional $60$ and $10$ epochs, respectively.
All experiments use the same internal multi-layer design dataset consisting of 4M high-quality samples.

\vspace{-1.5mm}
\paragraph{Evaluation setting.}
We mainly evaluate on \designbenchmark~\cite{pu2025art}\footnote{A graphic-design benchmark curated from VistaCreate~\cite{vistacreate}} and LICA~\cite{hirsch2026licalayeredimagecomposition}.
We report \emph{Layer} L1/PSNR/SSIM on non-transparent pixels using the ground-truth alpha mask, and \emph{Composite} L1/PSNR/SSIM between the composited prediction and the input image. 
We also report VLM-based ratings of \emph{Alpha} (matte and transparency quality) and \emph{Visual} (layer integrity and separation), judged by GPT-5.5~\cite{openai_gpt_5_5} and Gemini-3.1 Pro~\cite{google_gemini_3_1_pro}, and the system prompt is detailed in Supplementary~\ref{app:vlm_prompt}.
For the user study, we collect two-dimensional pairwise preferences from $10$ annotators with diverse backgrounds.

\vspace{-1.5mm}
\paragraph{Layout setting.}
To extract the regional tokens from flat design pixels, we adopt a graphic design layout detector DAD~\cite{dad2026anonymous} based on Qwen$3$-VL-$2$B~\cite{bai2025qwen3}.
The additional latency introduced by the layout detector is only $1$ second when predicting around $20$ bounding boxes in z-order.

\subsection{Main Results}

\vspace{-1.5mm}
\paragraph{Comparison with State-of-the-Art.}
We compare PixelART-H against state-of-the-art image-to-layer decomposers on \designbenchmark~\cite{pu2025art} and LICA~\cite{hirsch2026licalayeredimagecomposition}: Qwen-Image-Layered (QIL)~\cite{yin2025qwenimagelayered}, which adapts Qwen-Image-20B~\cite{wu2025qwenimage} with a VLD-MMDiT architecture, and CLD~\cite{cld}, an ART-style~\cite{pu2025art} variant built on FLUX.1[dev]-12B~\cite{flux2024}. Both methods rely on pretrained T2I backbones and RGBA latent autoencoders. We additionally include LayerD~\cite{suzuki2025layerd}, a non-diffusion pipeline based on matting and LaMa~\cite{suvorov2022resolution} inpainting. For fair comparison, we use the official checkpoint, codebase, and default inference configuration of each baseline. We use the same layout detector for PixelART and CLD, while QIL and LayerD are evaluated in their native bbox-free setting.

Table~\ref{tab:compare_main} shows that PixelART achieves both the best Composite PSNR and the highest VLM rating, despite using $6\times$ fewer parameters than QIL and $4\times$ fewer parameters than CLD. Beyond accuracy, PixelART-H also delivers substantial efficiency gains: measured on $100$ randomly sampled $1024\times1024$ inputs from \designbenchmark~\cite{pu2025art} on a single H200 GPU without any acceleration tricks (quantization, distillation, or step reduction), it runs $19\times$ faster than CLD and $69\times$ faster than QIL while using $8$--$11\times$ less peak GPU memory. These efficiency gains come from both the smaller diffusion transformer backbone and the absence of a VAE during sampling.\footnote{LayerD~\cite{suzuki2025layerd} reports a small parameter count but exhibits non-negligible latency due to costly CPU-side segmentation, matting, and inpainting operations in its official implementation.} 

\begin{figure}[t]
    \centering
    \includegraphics[width=\textwidth]{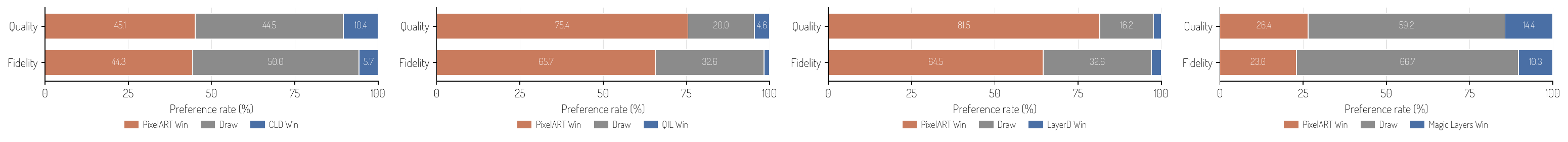}
    \vskip -2.5mm
    \caption{\small \textbf{User study.} Comparison with previous SoTA methods and the commercial system Magic Layers~\cite{magiclayers}.}
    \label{fig:user_study_compare}
    \vspace{-3mm}
\end{figure}

\begin{table}[t]
\centering
\footnotesize
\tablestyle{1pt}{1}
\resizebox{\linewidth}{!}{%
\begin{tabular}{l|ccc|ccc|ccc|cc}
    \multicolumn{4}{c|}{}
    & \multicolumn{3}{c|}{\designbenchmark}
    & \multicolumn{3}{c|}{LICA}
    & \multicolumn{2}{c}{} \\
    \cline{5-10}
    \multicolumn{4}{c|}{}
    & \multicolumn{1}{c|}{Pixel PSNR$\uparrow$} & \multicolumn{2}{c|}{VLM (GPT/Gem.)$\uparrow$}
    & \multicolumn{1}{c|}{Pixel PSNR$\uparrow$} & \multicolumn{2}{c|}{VLM (GPT/Gem.)$\uparrow$}
    & \multicolumn{2}{c}{Inference Efficiency} \\
    \cline{5-12}
    Method & T2I Pre. & VAE & {\makecell{\#Params (Layout)}}
    & Comp. & Alpha & Visual
    & Comp. & Alpha & Visual
    & Time (s)$\downarrow$ & Mem (GB)$\downarrow$ \\
    \shline
    CLD~\cite{cld} & \ding{51} & \ding{51} & 14B (12+2)
    & 29.79 & 4.24/4.83 & 4.50/4.68
    & 28.84 & 4.21/4.78 & 4.38/4.62
    & 58.7 (+0.7) & 55.6 (+4.2) \\
    Qwen-Image-Layered~\cite{yin2025qwenimagelayered} & \ding{51} & \ding{51} & 20B
    & 32.46 & 4.30/4.73 & 4.14/4.37
    & 31.99 & 4.22/4.65 & 3.93/4.20
    & 213.4 & 72.4 \\
    \rowcolor{gray!10}
    PixelART
    & \ding{55} & \ding{55} & 3.05B (1.05+2)
    & \textbf{36.92} & \textbf{4.32}/\textbf{4.91} & \textbf{4.57}/\textbf{4.81}
    & \textbf{36.39} & \textbf{4.26}/\textbf{4.87} & \textbf{4.44}/\textbf{4.75}
    & \textbf{3.1} (+0.7) & \textbf{6.6} (+4.2) \\
    \hline
    \graytext{LayerD$^\dagger$~\cite{suzuki2025layerd}} & \graytext{\ding{51}} & \graytext{\ding{51}} & \graytext{0.27B}
    & \graytext{36.07} & \graytext{4.12/3.94} & \graytext{3.49/3.13}
    & \graytext{37.22} & \graytext{4.06/4.01} & \graytext{3.64/3.37}
    & \graytext{15.0} & \graytext{3.4} \\
\end{tabular}%
}
\caption{\small \textbf{System-level comparison}. \textit{T2I Pre.}: whether initialized from pretrained T2I. \textit{Pixel PSNR}: PSNR between the alpha-composited reconstruction and the input image. \textit{VLM}: scores (1--5) from GPT-5.5 / Gemini-3.1 Pro (separated by `/') on alpha and visual quality. \textit{Efficiency}: per-image latency and peak memory on one H200. Parenthesized values for PixelART and CLD report the additional layout detector overhead (\#parameters, latency, GPU memory). LayerD$^\dagger$ is non-diffusion (segmentation + inpainting), which inflates Composite PSNR but yields the lowest VLM Visual score. Bold = best within the diffusion-based group.}
\label{tab:compare_main}
\vspace{-6mm}
\end{table}

\begin{figure}[t]
    \centering
    \includegraphics[width=0.9\textwidth]{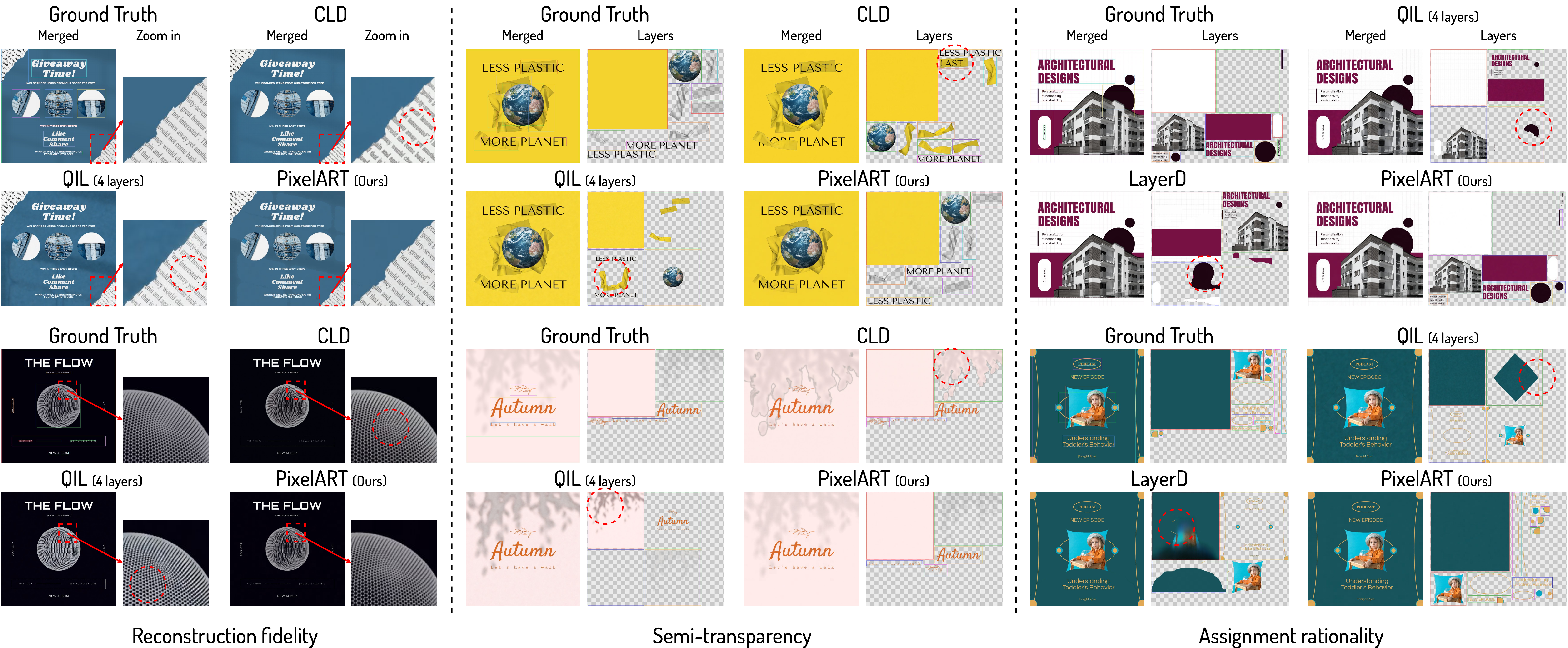}
    \caption{\small Qualitative comparison with state-of-the-art approaches. QIL: Qwen-Image-Layered.}
    \label{fig:transparent_text_compare}
    \vspace{-3mm}
\end{figure}

\vspace{-1.5mm}
\paragraph{Qualitative Comparison.}\label{sec:design_elements}
Figure~\ref{fig:transparent_text_compare} compares PixelART with prior methods along three complementary qualitative axes. (1) \textit{Reconstruction fidelity}: CLD and QIL often introduce visible artifacts on fine textures, small text, and fuse characters together, which is consistent with the information loss of latent autoencoding, whereas PixelART directly denoises RGBA pixels and better preserves high-frequency details. (2) \textit{Semi-transparency}: for soft shadows, feathered boundaries, and gradient overlays, CLD tends to collapse alpha into nearly binary masks and fails to produce transparency, while QIL bakes background colors into foreground layers and even introduces color shifts. In contrast, PixelART maintains significantly better RGB consistency and smooth alpha transitions. (3) \textit{Assignment rationality}: bbox-free baselines such as QIL and LayerD do not require layout input, but their outputs can mix unrelated elements, split coherent ones across layers, or suffer from segmentation leakage and incomplete extraction. On the other hand, PixelART uses bboxes as conditions, provides explicit control, and can produce more coherent editable layers.

\vspace{-1.5mm}
\paragraph{User Study.} Figure~\ref{fig:user_study_compare} summarizes the user study results. Our approach is consistently preferred over prior state-of-the-art methods and achieves comparable or superior performance to the latest commercial system, Canva Magic Layers~\cite{magiclayers}.

\vspace{-2mm}
\subsection{Ablation Study}

\vspace{-1.5mm}
\paragraph{Pixel-space outperforms latent-space diffusion for I2L.}
Table~\ref{tab:abl_pixel_vs_latent} compares pixel-space and latent-space diffusion under matched settings, differing only in whether layers are encoded through a pretrained RGBA-VAE. We use the RGBA-VAE from Qwen-Image-Layered~\cite{yin2025qwenimagelayered} as the strongest publicly available latent baseline. Pixel diffusion wins on every metric, with a +5.57 dB gain on Composite PSNR. These results indicate that latent representations are suboptimal for image-to-layer decomposition, and we therefore adopt pixel-space diffusion by default unless otherwise specified.

\begin{table}[!ht]
\begin{minipage}[t]{0.49\linewidth}
\centering
\tablestyle{6pt}{1}
\resizebox{\linewidth}{!}{%
\begin{tabular}{l|ccc|ccc}
    & \multicolumn{3}{c|}{Layer metrics} & \multicolumn{3}{c}{Composite metrics} \\
    Method & L1$\downarrow$ & PSNR$\uparrow$ & SSIM$\uparrow$ & L1$\downarrow$ & PSNR$\uparrow$ & SSIM$\uparrow$ \\
    \shline
    Latent VAE & 0.0749 & 26.37 & 0.8172 & 0.0202 & 28.79 & 0.9155 \\
    \rowcolor{gray!10}
   Pixel & \textbf{0.0626} & \textbf{28.66} & \textbf{0.8609} & \textbf{0.0132} & \textbf{34.36} & \textbf{0.9539} \\
\end{tabular}}
\captionof{table}{\small Latent diffusion vs. Pixel diffusion}
\label{tab:abl_pixel_vs_latent}
\end{minipage}
\hfill
\begin{minipage}[t]{0.49\linewidth}
\centering
\resizebox{\linewidth}{!}{%
\begin{tabular}{l|ccc|ccc}
    & \multicolumn{3}{c|}{Layer metrics} & \multicolumn{3}{c}{Composite metrics} \\
    Initialization & L1$\downarrow$ & PSNR$\uparrow$ & SSIM$\uparrow$ & L1$\downarrow$ & PSNR$\uparrow$ & SSIM$\uparrow$ \\
    \shline
    \rowcolor{gray!10}From scratch & \textbf{0.0626} & \textbf{28.66} & 0.8609 & \textbf{0.0132} & \textbf{34.36} & \textbf{0.9539} \\
    From T2I pretrained & 0.0639 & 28.27 & \textbf{0.8614} & 0.0132 & 33.97 & 0.9504 \\
\end{tabular}}
\captionof{table}{\small Effect of T2I pretraining.}
\label{tab:abl_t2i}
\end{minipage}

\begin{minipage}[t]{0.49\linewidth}
\centering
\footnotesize
\resizebox{\linewidth}{!}{
\begin{tabular}{l|ccc|ccc}
     & \multicolumn{3}{c|}{Layer metrics} & \multicolumn{3}{c}{Composite metrics} \\
    Prediction & L1$\downarrow$ & PSNR$\uparrow$ & SSIM$\uparrow$ & L1$\downarrow$ & PSNR$\uparrow$ & SSIM$\uparrow$ \\
    \shline
    \rowcolor{gray!10}$\mathbf{x}$-pred & \textbf{0.0626} & \textbf{28.66} & \textbf{0.8609} & \textbf{0.0132} & \textbf{34.36} & \textbf{0.9539} \\
    $\mathbf{v}$-pred & 0.2504 & 9.92 & 0.1718 & 0.2703 & 9.05 & 0.0383 \\
\end{tabular}}
\caption{\small Effect of prediction space.}
\label{tab:abl_pred_space}
\end{minipage}
\hfill
\begin{minipage}[t]{0.49\linewidth}
\centering
\resizebox{\linewidth}{!}{%
\begin{tabular}{l|ccc|ccc}
    & \multicolumn{3}{c|}{Layer metrics} & \multicolumn{3}{c}{Composite metrics} \\
    Optimizer & L1$\downarrow$ & PSNR$\uparrow$ & SSIM$\uparrow$ & L1$\downarrow$ & PSNR$\uparrow$ & SSIM$\uparrow$ \\
    \shline
    AdamW~\cite{loshchilov2019decoupled} & 0.0682 & 27.75 & 0.8475 & \textbf{0.0129} & 33.76 & 0.9501 \\
    \rowcolor{gray!10}
    Muon~\cite{jordan2024muon} & \textbf{0.0626} & \textbf{28.66} & \textbf{0.8609} & 0.0132 & \textbf{34.36} & \textbf{0.9539} \\
\end{tabular}}
\captionof{table}{\small Effect of optimizer.}
\label{tab:abl_optim}
\end{minipage}

\begin{minipage}[t]{0.49\linewidth}
\centering
\resizebox{\linewidth}{!}{%
\begin{tabular}{l|ccc|ccc}
    & \multicolumn{3}{c|}{Layer metrics} & \multicolumn{3}{c}{Composite metrics} \\
    Training data & L1$\downarrow$ & PSNR$\uparrow$ & SSIM$\uparrow$ & L1$\downarrow$ & PSNR$\uparrow$ & SSIM$\uparrow$ \\
    \shline
    40K & 0.0971 & 24.02 & 0.7671 & 0.0160 & 31.23 & 0.9292 \\
    400K & 0.0698 & 27.67 & 0.8424 & 0.0144 & 33.71 & 0.9479 \\
    \rowcolor{gray!10}
    4M  & \textbf{0.0626} & \textbf{28.66} & \textbf{0.8609} & \textbf{0.0132} & \textbf{34.36} & \textbf{0.9539} \\
\end{tabular}}
\captionof{table}{\small Effect of training data scale.}
\label{tab:abl_train_ds}
\end{minipage}
\hfill
\begin{minipage}[t]{0.49\linewidth}
\centering
\resizebox{\linewidth}{!}{%
\begin{tabular}{l|ccc|ccc}
    & \multicolumn{3}{c|}{Layer metrics} & \multicolumn{3}{c}{Composite metrics} \\
    Model variant & L1$\downarrow$ & PSNR$\uparrow$ & SSIM$\uparrow$ & L1$\downarrow$ & PSNR$\uparrow$ & SSIM$\uparrow$ \\
    \shline
    PixelART-B/16  & 0.0626 & 28.66 & 0.8609 & 0.0132 & 34.36 & 0.9539 \\
    PixelART-L/16  & 0.0475 & 31.78 & 0.8968 & \textbf{0.0094} & 36.79 & 0.9677 \\
    PixelART-H/16  & \textbf{0.0451} & \textbf{32.71} & \textbf{0.9011} & 0.0096 & \textbf{37.96} & \textbf{0.9696} \\
\end{tabular}}
\captionof{table}{\small Effect of model size scaling.}
\label{tab:abl_model_size}
\end{minipage}

\begin{minipage}[t]{1\linewidth}
\centering
\footnotesize
\tablestyle{15pt}{1}
\resizebox{\linewidth}{!}{
\begin{tabular}{l|cc|ccc|ccc}
    & \multicolumn{2}{c|}{Background metrics} & \multicolumn{3}{c|}{Layer metrics} & \multicolumn{3}{c}{Composite metrics} \\
    Loss & FID$\downarrow$ & LPIPS$\downarrow$ & L1$\downarrow$ & PSNR$\uparrow$ & SSIM$\uparrow$ & L1$\downarrow$ & PSNR$\uparrow$ & SSIM$\uparrow$ \\
    \shline
    MSE & 45.623 & 0.2910 & \textbf{0.0603} & \textbf{27.72} & \textbf{0.8431} & \textbf{0.0139} & \textbf{33.03} & \textbf{0.9410} \\
    + LPIPS (VGG) & \underline{41.478} & \underline{0.2739} & 0.0656 & 26.86 & 0.8324 & 0.0155 & 31.94 & 0.9314 \\
    \rowcolor{gray!10} + LPIPS (VGG, ConvNeXt) & \textbf{35.846} & \textbf{0.2577} & \underline{0.0631} & \underline{27.28} & \underline{0.8383} & \underline{0.0145} & \underline{32.46} & \underline{0.9360} \\
\end{tabular}
}
\caption{\small Effect of perceptual losses evaluated on the non-solid-background subset.}
\label{tab:abl_perceptual}
\end{minipage}

\begin{minipage}[t]{0.999\linewidth}
\centering
\footnotesize
\tablestyle{10pt}{1}
\resizebox{\linewidth}{!}{
\begin{tabular}{ll|ccc|ccc|ccc}
    & & \multicolumn{3}{c|}{Layer metrics} & \multicolumn{3}{c|}{Composite metrics} & \multicolumn{3}{c}{Assignment metrics} \\
    $q_{\text{boost}}$ & $p$ & L1$\downarrow$ & PSNR$\uparrow$ & SSIM$\uparrow$ & L1$\downarrow$ & PSNR$\uparrow$ & SSIM$\uparrow$ & $\alpha$-IoU$\uparrow$ & Owner Acc.$\uparrow$ & Leak.$\downarrow$ \\
    \shline
    -- & -- & 0.0626 & 28.66 & 0.8609 & 0.0132 & \underline{34.36} & 0.9539 & 0.8983 & 0.9040 & 0.1245 \\
    \hline
    \multirow{2}{*}{$\delta(t{=}0)$}        & 0.1 & 0.0588 & 27.89 & 0.8563 & 0.0137 & 33.06 & 0.9449 & 0.9471 & 0.9292 & 0.0987 \\
            & 0.2 & 0.0609 & 27.01 & 0.8444 & 0.0157 & 31.90 & 0.9327 & 0.9468 & 0.9347 & \underline{0.0943} \\
    \hline
    \multirow{2}{*}{$\mathcal{U}(0, 0.1)$}  & 0.1 & 0.0535 & \underline{28.89} & \underline{0.8778} & \underline{0.0125} & 34.08 & \underline{0.9545} & \underline{0.9476} & 0.9341 & 0.0951 \\
      & 0.2 & \underline{0.0523} & 28.87 & 0.8763 & 0.0125 & 33.82 & 0.9509 & \textbf{0.9501} & \textbf{0.9395} & \textbf{0.0909} \\
    \hline
    \multirow{2}{*}{$\mathcal{U}(0, 0.2)$}  & \hl 0.1 & \hl \textbf{0.0516} & \hl \textbf{29.51} & \hl \textbf{0.8802} & \hl \textbf{0.0119} & \hl \textbf{34.53} & \hl \textbf{0.9568} & \hl 0.9466 & \hl 0.9358 & \hl 0.0944 \\
      & 0.2 & 0.0534  & 28.86 & 0.8765 & 0.0125 & 34.01 & 0.9530 & 0.9476 & \underline{0.9377} & 0.0944 \\
\end{tabular}}
\caption{\small Effect of the boost component $q_{\text{boost}}$ and mixing weight $p$ in terminal-boosted timestep sampling.}
\label{tab:abl_time_dist}
\end{minipage}
\vspace{-7mm}
\end{table}

\begin{figure}[t]
    \centering
    \begin{minipage}[t]{0.68\linewidth}
        \vspace{0pt}
        \centering
        \subcaptionbox{\small Perceptual loss.\label{fig:percept_example}}{
            \includegraphics[width=\linewidth]{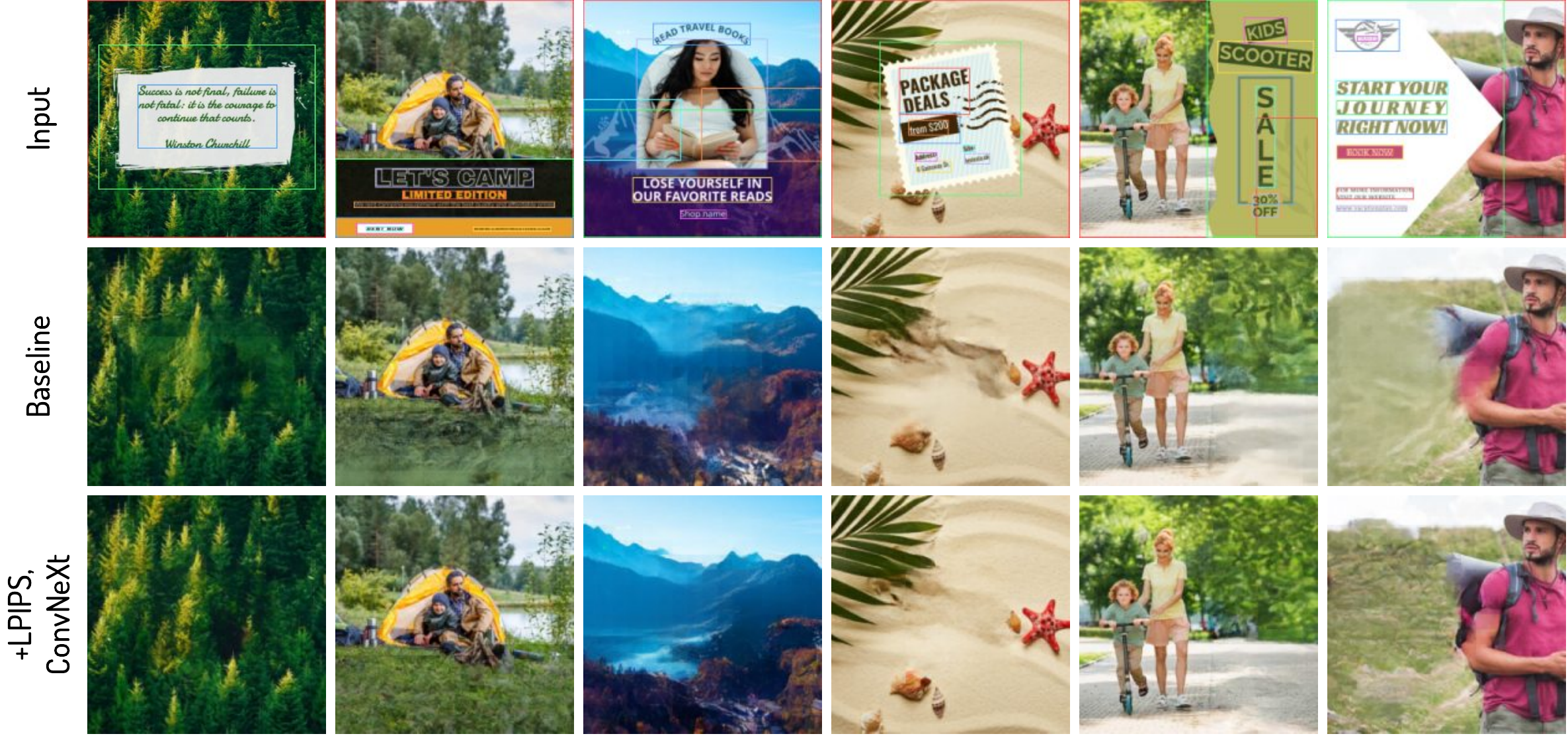}
        }
    \end{minipage}
    \hspace{3mm}
    \begin{minipage}[t]{0.225\linewidth}
        \vspace{0pt}
        \centering
        \subcaptionbox{\small Optimizer.\label{fig:abl_optim_curve}}{
            \includegraphics[width=\linewidth]{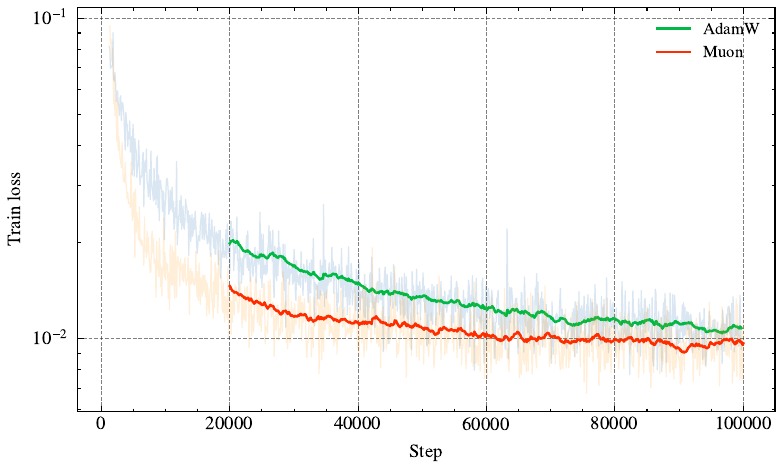}
        }

        \vspace{1mm}

        \subcaptionbox{\small T2I initialization.\label{fig:abl_t2i_loss}}{
            \includegraphics[width=\linewidth]{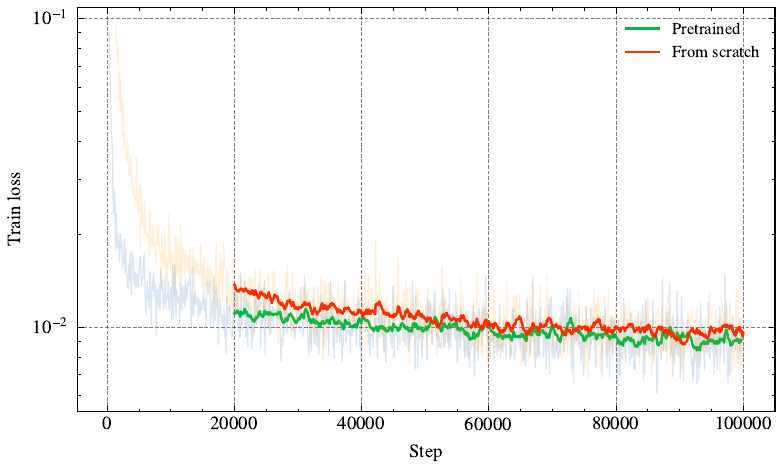}
        }
    \end{minipage}
    \vspace{-1mm}
    \caption{\small
    \textbf{Ablation visualizations.}
    \textbf{Left:} qualitative effect of perceptual losses on background reconstruction.
    \textbf{Right top:} training loss curves of Muon vs.\ AdamW under the same recipe.
    \textbf{Right bottom:} I2L fine-tuning loss curves for T2I-pretrained vs.\ from-scratch initializations on PixelART-B.
    }
    \label{fig:abl_visualizations}
    \vspace{-5mm}
\end{figure}

\vspace{-1.5mm}
\paragraph{T2I pretraining provides limited benefit for I2L.}
We test whether text-to-image pretraining improves image-to-layer decomposition. Following PixelGen~\cite{pixelgen}, we pretrain PixelART-B on the public BLIP-3o corpus~\cite{chen2025blip3o} ($\sim$36 million text--image pairs) for 250K steps with standard flow matching, then fine-tune on I2L using the same training budget as the from-scratch baseline. As shown in Table~\ref{tab:abl_t2i}, T2I pretraining brings a marginal improvement in the Layer SSIM metric ($+0.005$) and a slight regression on Composite PSNR ($-0.39$ dB). Figure~\ref{fig:abl_visualizations}(c) reveals the underlying training dynamics to better understand model behavior~\cite{liu2017towards}: the T2I-initialized model starts at a substantially lower loss thanks to its generic image generation prior, but the from-scratch run closes this gap within roughly 60K fine-tuning iterations. These results suggest that T2I priors transfer poorly to I2L decomposition once sufficient I2L training is applied. We hypothesize that this is because T2I models rely on text to steer denoising while I2L decomposition relies on the visible conditioning image, making the learned prior largely irrelevant.

\vspace{-1.5mm}
\paragraph{$\mathbf{x}$-prediction is critical for I2L.}
Table~\ref{tab:abl_pred_space} compares $\mathbf{x}$- and $\mathbf{v}$-prediction. $\mathbf{v}$-prediction fails to converge in our setting: each $16{\times}16{\times}4$ RGBA patch has a dimensionality of $1024$, exceeding the model's hidden width of $768$ and making it challenging to route a full-rank velocity target through the network. $\mathbf{x}$-prediction instead targets the clean image, which lies on a much lower-dimensional manifold of design content and fits comfortably within 768 channels, consistent with the manifold argument of JiT~\cite{li2025jit}. Even at PixelART-H scale with $d{=}1280$ that exceeds the patch dimensionality, $\mathbf{v}$-prediction still trails $\mathbf{x}$-prediction in Supplementary~\ref{app:pred_space} and Table~\ref{tab:supp_xv_h}.

\vspace{-1.5mm}
\paragraph{Muon outperforms AdamW.}
Table~\ref{tab:abl_optim} compares Muon~\cite{jordan2024muon} with AdamW~\cite{loshchilov2019decoupled} under matched learning rate, warmup, batch size, and training budget. Muon improves Layer PSNR by $+0.91$ dB and Composite PSNR by $+0.60$ dB, with consistent SSIM gains across both. Figure~\ref{fig:abl_visualizations}(b) further shows that Muon achieves a lower training loss under the same recipe.

\vspace{-1.5mm}
\paragraph{I2L benefits substantially from data scale.}
Table~\ref{tab:abl_train_ds} reports the effect of varying the training set across two orders of magnitude—from 40K to 4M real multi-layer designs—while holding all other settings fixed. Layer PSNR climbs by $+4.64$ dB (from $24.02$ to $28.66$) and Composite PSNR by $+3.13$ dB (from $31.23$ to $34.36$) over this range, with monotonic improvements on every metric. Although the marginal gain from $400$K to $4$M ($+0.99$ dB Layer PSNR) is smaller than from $40$K to $400$K ($+3.65$ dB), the curve has not yet plateaued, suggesting that further real-data scaling remains a promising direction. We adopt $4$M as our default training scale.

\vspace{-1.5mm}
\paragraph{PixelART scales monotonically with model size.}
Table~\ref{tab:abl_model_size} reports performance across three model variants—PixelART-B/16 ($168$M), PixelART-L/16 ($523$M), and PixelART-H/16 ($1.05$B)—trained with the same data and training budget. Layer PSNR climbs by $+4.05$ dB (from $28.66$ to $32.71$) and Composite PSNR by $+3.60$ dB (from $34.36$ to $37.96$) as the model grows from $168$M to $1.05$B parameters, with monotonic improvements on nearly every metric. Combined with the data-scaling results in Table~\ref{tab:abl_train_ds}, this confirms that PixelART scales effectively with both data and model size.

\vspace{-1.5mm}
\paragraph{Perceptual loss improves background reconstruction.}
Table~\ref{tab:abl_perceptual} ablates LPIPS-based perceptual supervision on the background layer, where large occluded regions make perceptual quality particularly visible to human observers. To avoid solid-color backgrounds dominating the evaluation, we report all metrics on the $1{,}474$-sample non-solid-background subset. Adding VGG~\cite{simonyan2015very} reduces background FID from $45.6$ to $41.5$ and LPIPS from $0.291$ to $0.274$; jointly using ConvNeXt~\cite{woo2023convnextv2} further improves them to $35.8$ and $0.258$, respectively, while Layer and Composite metrics remain comparable. Figure~\ref{fig:abl_visualizations}(a) confirms the trend qualitatively: perceptual supervision yields sharper and more coherent background completions. We therefore adopt both perceptual losses in the final system.

\vspace{-1.5mm}
\paragraph{Terminal-boosted sampling improves layer assignment.}
Table~\ref{tab:abl_time_dist} ablates the boost component $q_{\mathrm{boost}}$ and mixing probability $p$. Any boost---point-mass at $t{=}0$ or truncated-uniform $\mathcal{U}(0, \tau)$---improves all assignment metrics over the no-boost baseline, confirming that additional supervision in the high-noise regime helps the model learn layer assignment. Truncated-uniform boosts further outperform the point-mass variant on both reconstruction and assignment metrics.
Within truncated-uniform variants, we observe a consistent trade-off: stronger terminal weighting (larger $p$ or wider $\tau$) improves assignment metrics ($\alpha$-IoU, Owner Acc., Leakage) but slightly reduces Layer/Composite reconstruction quality. This matches the two-phase structure of I2L training: heavier terminal weighting allocates more capacity to the high-noise assignment phase at the expense of the low-noise refinement phase. We adopt $\mathcal{U}(0, 0.2)$ with $p{=}0.1$ as our default configuration, as it achieves the best reconstruction metrics while maintaining strong assignment performance.
\vspace{-2mm}
\section{Conclusion}
\label{sec:conclusion}
We introduce PixelART, a pixel-space rectified-flow Transformer trained from scratch for image-to-layer decomposition.
PixelART directly denoises regional RGBA pixel patches, removing the RGBA-VAE bottleneck and avoiding pretrained T2I backbones, and layer-specific decoders.
Our results indicate that layer decomposition is naturally modeled as a pixel-space assignment-and-completion problem: high-noise timesteps determine layer assignment, while low-noise timesteps refine RGBA details.
On \designbenchmark, PixelART achieves state-of-the-art layer decomposition and composite reconstruction with substantially lower latency and memory than VAE-based diffusion baselines.
Ablations show that pixel-space $\mathbf{x}$-prediction, high-noise timestep coverage, perceptual supervision, and data/model scaling are important, while T2I initialization provides marginal final gain in our data-rich I2L setting.
These findings establish from-scratch pixel-space diffusion as a simple and scalable approach for editable layer decomposition.



\bibliographystyle{IEEEtranN}
\bibliography{reference}


\newpage
\appendix

\section{Data Source}
\label{app:data_source}

\paragraph{Training Data.}
Our $4$M multi-layer training corpus is curated from internal multi-layer design templates that are explicitly licensed for model training. Each template is stored as an editable design file with vector and raster layers in z-order, and is rasterized at multiple resolutions ($256$, $512$, $1024$) into the \texttt{\{flat image, ordered RGBA layer stack, layer bounding boxes, layer order\}} tuples that PixelART consumes.

\paragraph{Evaluation Data.}
We use ~\designbenchmark~\cite{pu2025art}, which is curated from VistaCreate~\cite{vistacreate}, and LICA~\cite{hirsch2026licalayeredimagecomposition} as evaluation datasets; we do not include any of their test samples in training. No personally identifying information, copyrighted third-party images, or human-subject content were used, and no human annotators were employed for the data collection.

\section{Detailed Model and Training Configuration}
\label{app:config}
 
Table~\ref{tab:supp_config} summarizes the architecture, training, and sampling hyperparameters for all PixelART configurations used in the main paper. PixelART-B/L/H are the three variants used in the model-size ablation (PixelART-B is also the default for all other ablations); PixelART (main model) is a separately trained system-level model used for the comparison in Table~\ref{tab:compare_main}, sharing the -H scale but with a coarser patch size and a multi-resolution training curriculum. Hyperparameters are inherited from JiT~\cite{li2025jit} and SD3~\cite{esser2024sd3} defaults if not listed below.
 
\begin{table}[!ht]
\centering
\footnotesize
\caption{\textbf{Configurations of PixelART variants.} PixelART-B is the default for all ablations and is also the small model in the model-size ablation, alongside PixelART-L and PixelART-H. PixelART (main model) is the separately trained system-level model reported in Table~\ref{tab:compare_main}; it shares the H scale but uses a coarser patch size and a three-stage resolution curriculum. Settings that are common across multiple columns are merged. The pixel-space versus latent-space comparison (Table~\ref{tab:abl_pixel_vs_latent}) uses the same PixelART-B architecture, but the latter operates on the RGBA-VAE latents of Qwen-Image-Layered with a patch size of 2, with the parameter count differing by $\sim$1M due to the smaller patch size.}
\vspace{1mm}
\label{tab:supp_config}
\resizebox{\linewidth}{!}{%
\begin{tabular}{l|ccc|c}
\shline
 & PixelART-B & PixelART-L & PixelART-H & PixelART (main) \\
\shline
\multicolumn{5}{l}{\textbf{Architecture}} \\
\hline
\# Transformer blocks & 12 & 24 & 32 & 32 \\
hidden dim $d$ & 768 & 1024 & 1280 & 1280 \\
\# attention heads & 12 & 16 & 16 & 16 \\
patch size $P$ & \multicolumn{3}{c|}{16} & 32 \\
input noise scale & \multicolumn{3}{c|}{1} & 2 \\
patch-embed bottleneck dim $d_b$ & 128 & 128 & 256 & 256 \\
\# parameters & 168M & 523M & 1052M & 1057M \\
MLP expansion ratio & \multicolumn{4}{c}{4} \\
positional encoding & \multicolumn{4}{c}{3D RoPE over $(x, y, z)$} \\
\hline
\multicolumn{5}{l}{\textbf{Training}} \\
\hline
prediction target & \multicolumn{4}{c}{$\mathbf{x}$-prediction with $\mathbf{v}$-loss} \\
base timestep schedule & \multicolumn{4}{c}{$\mathrm{logit}(t)\sim\mathcal N(\mu{=}{-}0.8,\sigma{=}0.8)$} \\
optimizer & \multicolumn{4}{c}{Muon~\cite{jordan2024muon}} \\
learning rate & \multicolumn{4}{c}{$1\mathrm{e}{-}4$ (constant)} \\
warmup steps & \multicolumn{4}{c}{1000} \\
weight decay & \multicolumn{4}{c}{0} \\
EMA decay & \multicolumn{4}{c}{0.9996} \\
text-condition drop (CFG) & \multicolumn{4}{c}{0.1} \\
boost component $q_{\mathrm{boost}}$ & \multicolumn{3}{c|}{not used} & $\mathcal U(0, 0.2)$ with $p{=}0.1$ \\
perceptual loss (at $t>0.3$) & \multicolumn{3}{c|}{not used} & LPIPS (VGG, ConvNeXt-V2) \\
hardware & \multicolumn{3}{c|}{$8\times$ H200} & $32\times$ H200 \\
training resolution & \multicolumn{3}{c|}{$256$} & $256\!\to\!512\!\to\!1024$ \\
batch size (per resolution) & \multicolumn{3}{c|}{$2048$} & $2048/2048/1024$ \\
training duration (per resolution) & \multicolumn{3}{c|}{$100$K steps ($\sim$20 epochs)} & $80/60/10$ epochs \\
\hline
\multicolumn{5}{l}{\textbf{Sampling}} \\
\hline
inference time shift & \multicolumn{3}{c|}{not used} & $3$ (at $1024\,$px only) \\
ODE solver & \multicolumn{4}{c}{Euler} \\
\# sampling steps & \multicolumn{4}{c}{25} \\
timestep grid & \multicolumn{4}{c}{linear in $[0, 1]$} \\
CFG scale & \multicolumn{4}{c}{$2.2$} \\
\shline
\end{tabular}}
\end{table}

\section{Layout Detector}
\label{app:layout_detector}

PixelART is a region-conditioned image-to-layer model: given a z-ordered set of
regional bounding boxes $\mathcal R$, it predicts an editable RGBA crop for each
region. During training, $\mathcal R$ is obtained from the ground-truth design
file. At inference time, $\mathcal R$ can either be specified by users to control
the desired layer granularity, or be predicted by a layout detector from the flattened
RGB image to enable fully automatic image-to-layer decomposition.

For system-level comparisons, we use DAD~\cite{dad2026anonymous}, an internal
graphic-design layout detector based on Qwen$3$-VL-$2$B~\cite{bai2025qwen3}.
The detector predicts foreground bounding boxes in back-to-front z-order.
PixelART only uses the predicted box coordinates and their order; no masks,
alpha mattes, RGB content, semantic labels, or layer pixels are passed from the
detector. Thus, the layout detector serves as an automatic interface for
providing the structural input required by PixelART.

The detector is trained on internal design data disjoint from the
\designbenchmark and LICA test splits. On the $1024\times1024$ \designbenchmark
inputs reported in Table~\ref{tab:compare_main}, it adds $0.7$~s latency and
$4.2$~GB peak GPU memory per image, reported separately in the parenthesized
terms of Table~\ref{tab:compare_main}.

\section{Evaluation Metric Definitions}
\label{app:metrics}
 
We evaluate three groups of pixel-level metrics: \emph{Layer} metrics for decomposition fidelity, \emph{Composite} metrics for visible reconstruction fidelity, and \emph{Assignment} metrics for layer-assignment correctness. Throughout, let $\hat{\mathbf L}_i\in[0,1]^{H\times W\times 4}$ and $\mathbf L_i\in[0,1]^{H\times W\times 4}$ denote the predicted and ground-truth RGBA layer $i$, respectively, with $\mathbf L_i = (\mathbf L_i^{\mathrm{rgb}}, \alpha_i)$. We use $\mathbf I, \hat{\mathbf I}\in[0,1]^{H\times W\times 3}$ for the input flat image and the predicted alpha-composite, and $K{+}1$ for the total number of layers (background plus $K$ foreground).
 
\subsection{Layer Metrics}
Layer metrics measure per-layer reconstruction fidelity. To jointly capture errors in both the predicted alpha and the predicted RGB channels, we compare alpha-premultiplied color, $\hat\alpha_i(p)\,\hat{\mathbf L}_i^{\mathrm{rgb}}(p)$ versus $\alpha_i(p)\,\mathbf L_i^{\mathrm{rgb}}(p)$, with the per-pixel error further weighted by the ground-truth alpha. A pixel for which the predicted alpha is wrong then contributes a non-zero error even if its RGB happens to match. Concretely, for each layer $i$ and pixel $p$, the layer L1 is
\begin{equation}
\mathrm{L1}_{\mathrm{layer}}^{(i)}
=
\frac{\sum_{p}\alpha_i(p)\,\big\|\hat\alpha_i(p)\,\hat{\mathbf L}_i^{\mathrm{rgb}}(p)\,-\,\alpha_i(p)\,\mathbf L_i^{\mathrm{rgb}}(p)\big\|_1}
     {3\sum_{p}\alpha_i(p)+\varepsilon}.
\end{equation}
PSNR and SSIM are computed analogously on the same alpha-premultiplied RGB pair, weighted by $\alpha_i$. We then average over all layers in the test sample, and finally over the test set.
 
\subsection{Composite Metrics}
Composite metrics measure visible reconstruction fidelity. We first composite the predicted layers in z-order:
\begin{equation}
\hat{\mathbf I}
=
\Pi_{\mathrm{RGB}}\left(\operatorname{Over}(\hat{\mathbf L}_0, \hat{\mathbf L}_1, \ldots, \hat{\mathbf L}_K)\right),
\end{equation}
and report L1, PSNR, and SSIM between $\hat{\mathbf I}$ and the input flat image $\mathbf I$ over all pixels.
 
\subsection{Layer Assignment Metrics}
\label{app:assignment_metrics}
Pixel reconstruction metrics alone do not penalize a model for confusing two visually similar layers, so we additionally report three layer-assignment metrics. Let $\hat\alpha_i, \alpha_i\in[0,1]^{H\times W}$ denote the predicted and ground-truth alpha of layer $i$, with index $i=0$ for the background and $i\ge 1$ for foreground layers in increasing $z$-order.
 
\paragraph{Alpha-IoU (soft IoU on alpha masks).}
We measure how well the predicted alpha masks overlap with the ground-truth ones via a continuous (soft) intersection-over-union, computed independently per layer and averaged across all layers including the background:
\begin{equation}
\alpha\text{-IoU}_i
=
\frac{\sum_p \min\!\big(\hat\alpha_i(p),\,\alpha_i(p)\big)}
     {\sum_p \max\!\big(\hat\alpha_i(p),\,\alpha_i(p)\big) + \varepsilon},
\qquad
\alpha\text{-IoU}
=
\frac{1}{K+1}\sum_{i=0}^{K}\alpha\text{-IoU}_i.
\label{eq:alpha_iou}
\end{equation}
The min/max formulation reduces to the standard IoU when $\hat\alpha_i, \alpha_i\in\{0,1\}$ and remains well-defined for soft alpha. Higher is better.
 
\paragraph{Visible-pixel ownership (Owner Acc.).}
Owner accuracy answers that for each pixel the user actually sees, whether the model assigns it to the right layer.
For a layer stack with index $i$ increasing toward the front, the visible contribution of layer $i$ at pixel $p$ after $z$-ordered source-over compositing is
\begin{equation}
w_i(p) \;=\; \alpha_i(p)\!\!\prod_{k>i}\!\big(1-\alpha_k(p)\big),
\label{eq:visible_weight}
\end{equation}
and the per-pixel owner is the layer with the largest visible contribution:
\begin{equation}
o^{\mathrm{gt}}(p) = \arg\max_i w_i^{\mathrm{gt}}(p),
\qquad
o^{\mathrm{pred}}(p) = \arg\max_i w_i^{\mathrm{pred}}(p).
\end{equation}
We restrict the metric to foreground pixels with a clearly dominant ground-truth owner: $\Omega_{\mathrm{fg}}=\{p:o^{\mathrm{gt}}(p)>0,\;\max_i w_i^{\mathrm{gt}}(p)>\tau_{\mathrm{vis}},\;w_{(1)}^{\mathrm{gt}}(p)-w_{(2)}^{\mathrm{gt}}(p)>\delta\}$, where $w_{(1)},w_{(2)}$ are the top-1 and top-2 visible weights and we use $\tau_{\mathrm{vis}}=\delta=0.05$. The background layer $i=0$ is excluded from $\Omega_{\mathrm{fg}}$ to prevent it from dominating the metric. Owner accuracy is the fraction of $\Omega_{\mathrm{fg}}$ for which the predicted and ground-truth owners agree:
\begin{equation}
\mathrm{Owner\text{-}Acc.}
=
\frac{1}{|\Omega_{\mathrm{fg}}|}\sum_{p\in\Omega_{\mathrm{fg}}}\mathbf{1}\!\big[o^{\mathrm{pred}}(p) = o^{\mathrm{gt}}(p)\big].
\label{eq:owner_acc}
\end{equation}
 
\paragraph{Wrong-layer leakage (Leak.).}
Wrong-layer leakage complements Owner accuracy by measuring \emph{how much} predicted mass is misplaced rather than only whether the top-1 owner is correct. On each ground-truth owner mask $M_i^{\mathrm{gt}}=\{p:o^{\mathrm{gt}}(p)=i\}$, it reports the fraction of predicted visible contribution that lands on layers other than $i$. We use the visible-weight form $w_j^{\mathrm{pred}}$ rather than raw alpha so that the metric remains stable in regions with semi-transparent overlays such as drop shadows, glows, and gradient fills:
\begin{equation}
\mathrm{Leak.}
=
\frac{\sum_i \sum_{p\in M_i^{\mathrm{gt}}} \sum_{j\neq i} w_j^{\mathrm{pred}}(p)}
     {\sum_i \sum_{p\in M_i^{\mathrm{gt}}} \sum_{j} w_j^{\mathrm{pred}}(p) + \varepsilon}.
\label{eq:leakage}
\end{equation}
Lower is better. The denominator normalizes by the total predicted visible mass on the same set of pixels, so the metric is a unitless leakage ratio independent of overall alpha calibration.
 
\subsection{VLM Quality Metrics}
\label{app:vlm_prompt}
For each test sample, we render a single PNG with the original input on the left and a packed grid of the predicted RGBA layers on the right (each cell labeled \texttt{L1}, \texttt{L2}, \dots, in $z$-order), and ask a frontier VLM to score every layer along two dimensions on a $1$--$5$ scale: \emph{Alpha} (matte and edge quality, opacity correctness) and \emph{Visual} (content integrity, object completeness, clean foreground/background separation). The exact prompt is shown in the listing below; it is identical across all methods and datasets compared in Table~\ref{tab:compare_main}, with only the candidate decomposition image differing. The VLM returns a strict JSON record with one Alpha and one Visual score per layer. We parse this record, average per sample, and then average across the test set. We report results from GPT-5.5~\cite{openai_gpt_5_5} (with \textit{high} image detail and default reasoning effort) and Gemini-3.1 Pro~\cite{google_gemini_3_1_pro} (with \textit{high} media resolution and default thinking level) separately.
 
\begin{tcolorbox}[breakable, colback=gray!5, colframe=black!50, title={\bfseries VLM judge prompt for layer-decomposition quality.}, fonttitle=\bfseries, fontupper=\footnotesize]
You are an expert judge for layer decomposition quality. You will evaluate \textbf{all layers shown in a single concatenated image}.
 
\medskip
\textbf{\#\# Image Layout}
 
\smallskip
The input image has this structure:
\begin{itemize}[leftmargin=1.5em, itemsep=0pt, topsep=2pt]
\item \textbf{Left side}: The original composite image (reference).
\item \textbf{Right side}: A grid of layer cutouts (collage layout). Each cell is labeled \texttt{LN} (e.g., \texttt{L1}, \texttt{L2}, \dots) at the bottom. Larger elements occupy bigger cells.
\end{itemize}
 
\medskip
\textbf{\#\# Your Task}
 
\smallskip
Score \textbf{ALL} the layers shown in the grid, using \textbf{two dimensions} (1--5, higher is better).
 
\smallskip
\textbf{Score Scale}:
\begin{itemize}[leftmargin=1.5em, itemsep=0pt, topsep=2pt]
\item \textbf{5} = Excellent, no visible issues.
\item \textbf{4} = Good with minor issues.
\item \textbf{3} = Acceptable with noticeable problems.
\item \textbf{2} = Poor, significant problems.
\item \textbf{1} = Unusable, severe failures.
\end{itemize}
 
\medskip\hrule\medskip
 
\textbf{\#\#\# Dimension 1: Alpha Quality (Transparency \& Edge Integrity)}
 
\smallskip
Evaluate the \textbf{matte quality}, edge precision, and transparency correctness. This dimension focuses on the alpha channel and edge processing.
 
\smallskip
\textbf{What to check}:
\begin{itemize}[leftmargin=1.5em, itemsep=0pt, topsep=2pt]
\item Edge artifacts: Haloing (light/dark outline), fringing (color bleeding at edges), jagged/aliased edges.
\item Opacity correctness: Is transparency logical? (glass = semi-transparent, solid objects = opaque).
\item Processing artifacts: Grid patterns in semi-transparent areas, alpha channel noise.
\end{itemize}
 
\smallskip
\textbf{Score Scale}:
\begin{itemize}[leftmargin=1.5em, itemsep=0pt, topsep=2pt]
\item \textbf{5}: Excellent matte. Crisp edges, smooth transitions, no halos or artifacts.
\item \textbf{4}: Good. Minor edge softness or faint halo, clean overall.
\item \textbf{3}: Acceptable. Visible halo, aliasing, or minor grid artifacts.
\item \textbf{2}: Poor. Strong jagged edges, heavy halos, distinct grid patterns.
\item \textbf{1}: Unusable. Massive halos, wrong transparency, severe alpha failure.
\end{itemize}
 
\medskip\hrule\medskip
 
\textbf{\#\#\# Dimension 2: Visual Quality (Content Integrity \& Segmentation)}
 
\smallskip
Evaluate the pixel integrity, semantic correctness, \textbf{completeness}, and \textbf{clean separation} of the visible content.
 
\smallskip
\textbf{CRITICAL --- Object Completeness Check}: For EACH layer, carefully compare it with the original image and ask:
\begin{enumerate}[leftmargin=1.5em, itemsep=0pt, topsep=2pt]
\item \textbf{Is this a complete semantic object?} Look at the original --- if you can see a whole surfboard, person, logo, etc., the layer should contain the ENTIRE object, not just a piece of it.
\item \textbf{Is the layer cleanly separated?} Background layers should NOT contain fragments of foreground objects.
\end{enumerate}
 
\smallskip
\textbf{What to check}:
\begin{itemize}[leftmargin=1.5em, itemsep=0pt, topsep=2pt]
\item Visual artifacts: Strange gray values, noise, compression artifacts, color bleeding.
\item Structural issues: Distorted anatomy (extra fingers, melted faces), garbled text, twisted shapes.
\item Content fidelity: Does it match the corresponding element in the original?
\end{itemize}
 
\smallskip
\textbf{Segmentation Failures} (penalize these severely):
\begin{itemize}[leftmargin=1.5em, itemsep=0pt, topsep=2pt]
\item \textbf{Incomplete object}: The layer contains only PART of an object that is COMPLETE in the original. Example: Original shows a full surfboard $\rightarrow$ Layer shows only half of it. This is WRONG --- the entire surfboard should be in one layer.
\item \textbf{Foreground residue}: Background layers contain leftover pieces of foreground objects. Example: A background layer has someone's arm or part of an object floating in it.
\item \textbf{Missing inpainting}: An object is partially hidden behind another object in the original, but the layer doesn't complete the hidden portion.
\end{itemize}
 
\smallskip
\textbf{Note on image boundary}: If an object is cut off at the IMAGE BOUNDARY in the original (e.g., half a person at the edge of the photo), the layer cannot show the missing part because the output layer has the same dimensions as the original --- there's simply no canvas space beyond the boundary. This is NOT a failure. However, if an object is fully WITHIN the original image, it must be fully present in the layer.
 
\smallskip
\textbf{Score Scale}:
\begin{itemize}[leftmargin=1.5em, itemsep=0pt, topsep=2pt]
\item \textbf{5}: Excellent. All objects are complete, clean separation, no artifacts.
\item \textbf{4}: Good. Minor artifacts. All objects complete and cleanly separated.
\item \textbf{3}: Acceptable. Some artifacts. Minor incompleteness or small residue.
\item \textbf{2}: Poor. \textbf{Object is incomplete} (cut off mid-object when it is complete in the original), OR noticeable foreground residue on the background layer.
\item \textbf{1}: Unusable. Severe incompleteness, major residue, or unrecognizable content.
\end{itemize}
 
\medskip\hrule\medskip
 
\textbf{\#\# Output Requirements}
 
\smallskip
A) Return \textbf{strict JSON only} (no markdown fences, no extra text). \\
B) Output scores for \textbf{ALL} layers visible in the grid (\texttt{L1}, \texttt{L2}, \texttt{L3}, \dots). \\
C) Use concise \textbf{failure\_tags}:
\begin{itemize}[leftmargin=1.5em, itemsep=0pt, topsep=2pt]
\item \emph{Alpha}: \texttt{["halo", "jagged", "fringing", "grid\_artifacts", "wrong\_opacity", "edge\_noise"]}.
\item \emph{Visual}: \texttt{["distorted\_anatomy", "bad\_text", "noise", "color\_bleed", "blur", "pixel\_corruption", "incomplete\_object", "foreground\_residue", "missing\_inpainting", "abnormal\_gray"]}.
\item \texttt{incomplete\_object} = \textbf{IMPORTANT}: object is complete in the original but layer only shows a fragment.
\item \texttt{foreground\_residue} = the background layer has leftover pieces of foreground objects (not cleanly removed).
\item \texttt{missing\_inpainting} = hidden/occluded parts are not filled in.
\end{itemize}
D) Provide a short, specific \textbf{reason} for each score.
 
\medskip
\textbf{\#\# Output JSON Schema}
\begin{verbatim}
{
  "layer_count": <number of layers evaluated>,
  "layers": [
    {
      "layer_id": "L1",
      "alpha_quality": {
        "score": 1-5,
        "failure_tags": [],
        "reason": "string"
      },
      "visual_quality": {
        "score": 1-5,
        "failure_tags": [],
        "reason": "string"
      }
    }
  ],
  "overall_summary": "string"
}
\end{verbatim}
Output strict JSON only.
\end{tcolorbox}
 
We invoke the VLM in a single round per sample. Responses are validated against the JSON schema above; samples whose response does not parse are re-queried up to three times before being discarded.

\section{User Study Protocol}
\label{app:user_study}
 
The user study referenced in the main text complements the VLM-based metrics with pairwise human preferences. We describe recruitment and informed consent, the evaluation interface and protocol, and data-handling procedures.
 
\paragraph{Recruitment and informed consent.} We recruit 10 annotators with diverse backgrounds. Before participation, each annotator receives a written informed-consent document covering: (i) the purpose of the study (comparing layer-decomposition methods on layer quality and composite fidelity); (ii) the procedure and expected duration (approximately $20$--$40$ minutes); (iii) the data we collect (a self-chosen username, per-question choices, optional failure-mode tags, and optional free-text notes; no audio, video, or sensitive personal information); (iv) voluntary participation and the right to withdraw at any time, including post-hoc deletion of submitted data on request; and (v) the intended use of the results (anonymized, aggregated reporting). Annotators sign the document before beginning the task.
 
\begin{figure}[!ht]
\centering
\includegraphics[width=\linewidth]{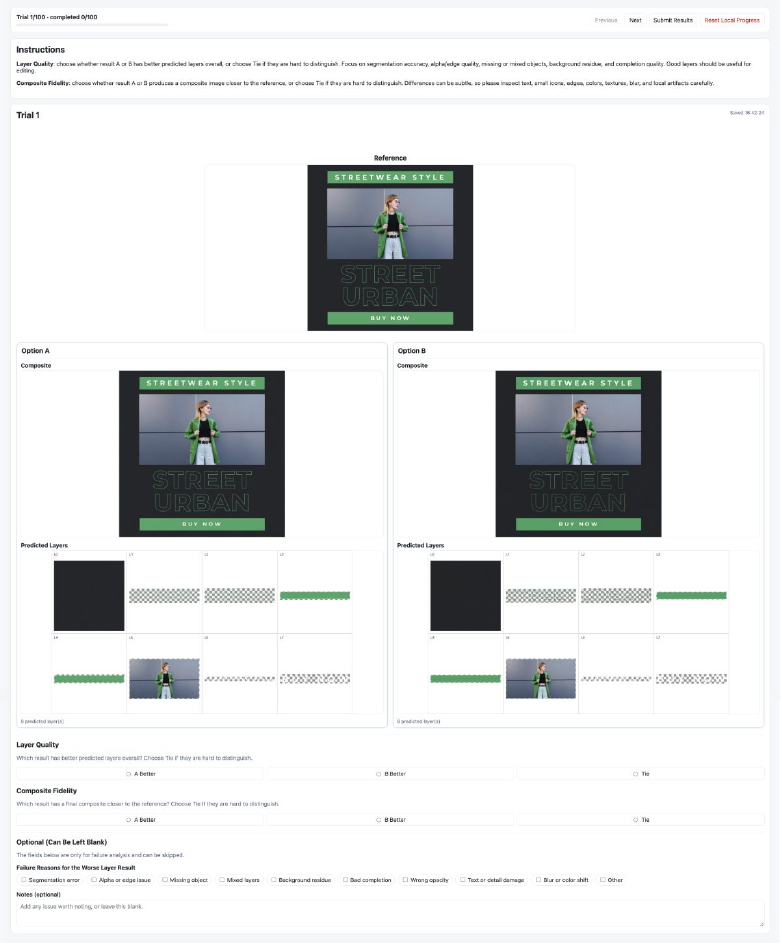}
\caption{\textbf{Web interface used in the user study.} Each question shows the original input image, two anonymized layer-decomposition outputs (A/B), the two-dimension ternary choice, and optional fields for failure tags and free-text notes.}
\label{fig:user_study_interface}
\end{figure}

\paragraph{Evaluation interface and protocol.} Annotators complete the study through a web-based questionnaire. Each question presents the original input image alongside two anonymized layer-decomposition outputs labelled A and B, produced by two different methods. For each pair, the annotator independently selects one of three options---``A is better'', ``B is better'', or ``tie''---on each of two dimensions:
\begin{itemize}[leftmargin=2em, itemsep=2pt, topsep=2pt]
\item \textbf{Layer quality}: how suitable the predicted layers are for downstream editing, including segmentation accuracy, alpha and edge quality, missing or mixed objects, residual foreground content in the background layer, and plausibility of inpainted regions.
\item \textbf{Composite consistency}: how faithfully the alpha-composited reconstruction matches the input, including text, small icons, edges, colour and texture, blur, and localized artefacts.
\end{itemize}
The annotator may optionally add failure-mode tags and free-text notes; these fields are not required. Method identities (the A/B assignment) are randomized per question and not disclosed during the study. Figure~\ref{fig:user_study_interface} shows the annotation interface.
 
\paragraph{Data handling and privacy.} Submitted responses are stored on internal infrastructure accessible only to the research team. Beyond the self-chosen username we do not collect age, contact information, or any identifying data, and annotators are instructed not to enter sensitive personal information in free-text notes. We report only aggregated, anonymized statistics; raw responses are retained internally and are deleted on participant request.

\section{Prediction Space Discussion}
\label{app:pred_space}
 
This section discusses the prediction space of PixelART. We first explain why our $\mathbf{x}$-prediction $\mathbf{v}$-loss formulation is compatible with terminal-boosted timestep sampling (Sec.~\ref{sec:terminal_boost}), and then complement the main-text ablation (Table~\ref{tab:abl_pred_space}) with $\mathbf{x}$- vs.\ $\mathbf{v}$-prediction at the PixelART-H scale.
 
\subsection{Compatibility with \texorpdfstring{$\mathbf{x}$}{x}-prediction \texorpdfstring{$\mathbf{v}$}{v}-loss}
\label{app:pred_compat}
The boost distribution must avoid the clean-data endpoint. Because our $\mathbf{x}$-prediction $\mathbf{v}$-loss has an equivalent clean-prediction weight $1/(1-t)^2$, any boost density with mass arbitrarily close to $t=1$ would make the expected loss unstable. Our candidates, $\delta(t=0)$ and $\mathcal U(0,\tau)$ with $\tau<1$, are bounded away from $t=1$, so
\begin{equation}
\mathbb E_{t\sim q_{\mathrm{boost}}}\!\left[\frac{1}{(1-t)^2}\,\big\|\widehat{\mathbf Y}_{\theta}-\mathbf Y\big\|_2^2\right]<\infty.
\end{equation}
For $\mathcal U(0,\tau)$, the weight is upper-bounded by $1/(1-\tau)^2$.
 
\subsection{PixelART-H \texorpdfstring{$\mathbf{x}$}{x}- vs.\ \texorpdfstring{$\mathbf{v}$}{v}-Prediction}
\label{app:xvpred_h}
The main-text ablation (Table~\ref{tab:abl_pred_space}) shows that $\mathbf{v}$-prediction fails to converge at the PixelART-B scale, where each $16{\times}16{\times}4$ RGBA patch has dimensionality $1024$ exceeding the model's hidden width of $768$. To test whether the failure is purely a capacity issue, we repeat the comparison at the PixelART-H scale ($d{=}1280$, exceeding the patch dimensionality) with the convolutional bottleneck removed, so the per-token dimensionality entering the Transformer is $1024$ rather than the bottleneck width. The training recipe otherwise follows the PixelART-H ablation default in Table~\ref{tab:supp_config}.
 
\begin{table}[!ht]
\centering
\footnotesize
\caption{\textbf{$\mathbf{x}$- vs.\ $\mathbf{v}$-prediction at PixelART-H scale} ($d{=}1280$, no bottleneck). Both runs use identical recipes apart from the prediction target. At this width, $\mathbf{v}$-prediction \emph{does} converge but still trails $\mathbf{x}$-prediction, indicating that the manifold mismatch is not solely a capacity issue.}
\label{tab:supp_xv_h}
\begin{tabular}{l|ccc|ccc}
& \multicolumn{3}{c|}{Layer metrics} & \multicolumn{3}{c}{Composite metrics} \\
Prediction & L1$\downarrow$ & PSNR$\uparrow$ & SSIM$\uparrow$ & L1$\downarrow$ & PSNR$\uparrow$ & SSIM$\uparrow$ \\
\shline
\rowcolor{gray!10}
$\mathbf{x}$-pred & \textbf{0.0453} & \textbf{32.88} & \textbf{0.8995} & \textbf{0.0086} & \textbf{38.55} & \textbf{0.9727} \\
$\mathbf{v}$-pred & 0.0526 & 30.17 & 0.8691 & 0.0114 & 36.08 & 0.9441 \\
\end{tabular}
\end{table}
 
At PixelART-H, $\mathbf{v}$-prediction trails $\mathbf{x}$-prediction by $-2.71$~dB Layer PSNR and $-2.47$~dB Composite PSNR. This residual gap is consistent with the manifold argument of JiT~\cite{li2025jit}: design layers occupy a much lower-dimensional manifold than the velocity target, and predicting clean pixels remains a better-conditioned regression problem than predicting velocities, even at high width.

\section{RGBA-VAE Decoder for the Latent-Space Baseline}
\label{app:vae_decoder}
 
The pixel-space versus latent-space comparison (Table~\ref{tab:abl_pixel_vs_latent}) trains a PixelART-B variant on the RGBA-VAE latents released with Qwen-Image-Layered~\cite{yin2025qwenimagelayered}, using a patch size of $2$ to match the latent's spatial down-sampling; all other settings follow the PixelART-B ablation default in Table~\ref{tab:supp_config}. PixelART operates on bounding-box-conditioned regional latents, but the official Qwen-Image-Layered decoder reconstructs only full-canvas latents and cannot be applied directly. We therefore train a separate ViT-based RGBA decoder (300M parameters) on the same 4M design-template corpus as PixelART, with the original encoder kept frozen; this decoder maps cropped regional latents to cropped RGBA layers, which are then composited into the canvas.
 
\begin{table}[!ht]
\centering
\footnotesize
\caption{\textbf{RGBA-VAE decoder reconstruction quality} on \designbenchmark{} at $256\times256$. Each decoder takes the official Qwen-Image-Layered encoder's latent of a ground-truth RGBA layer and produces a reconstruction, which is compared against the original. Both decoders share the same encoder. Layer L1/PSNR are alpha-pre-multiplied (Appendix~\ref{app:metrics}); layer SSIM is alpha-weighted.}
\label{tab:supp_vae_decoder}
\begin{tabular}{l|ccc|ccc}
& \multicolumn{3}{c|}{Composite metrics} & \multicolumn{3}{c}{Layer metrics} \\
Decoder & L1$\downarrow$ & PSNR$\uparrow$ & SSIM$\uparrow$ & L1$\downarrow$ & PSNR$\uparrow$ & SSIM$\uparrow$ \\
\shline
Qwen-Image-Layered (official) & 0.0132 & 32.52 & 0.9568 & 0.0363 & 31.66 & 0.9289 \\
\rowcolor{gray!10}
Ours (300M ViT) & \textbf{0.0083} & \textbf{33.69} & \textbf{0.9686} & \textbf{0.0254} & \textbf{37.49} & \textbf{0.9569} \\
\end{tabular}
\end{table}
 
Table~\ref{tab:supp_vae_decoder} reports reconstruction metrics for our trained decoder and the official Qwen-Image-Layered decoder under the same encoder. Our decoder is competitive with the official one on every metric, providing a comparable upper bound for any latent diffusion model that uses this encoder.

\section{T2I-Pretrained PixelART-B Performance}
\label{app:t2i_pretrain}
 
For the T2I-pretraining ablation in Table~\ref{tab:abl_t2i}, we pretrain PixelART-B (168M parameters) on the public BLIP-3o corpus~\cite{chen2025blip3o} ($\sim$36M text--image pairs) for 250K steps with standard rectified-flow training, then fine-tune on our I2L corpus with the same recipe as the from-scratch baseline. Unless stated otherwise, all training hyperparameters follow the PixelART-B ablation default in Table~\ref{tab:supp_config}. To verify that the pretrained checkpoint produces coherent T2I outputs, we evaluate it on GenEval~\cite{ghosh2023geneval} \emph{before} I2L fine-tuning. The full breakdown is:
 
\begin{table}[!ht]
\centering
\footnotesize
\caption{\textbf{GenEval scores of the BLIP-3o-pretrained PixelART-B} (168M params, batch 1024, 250K iters), evaluated before I2L fine-tuning. Overall is the per-task average. The model does not adopt any specialized optimization recipes common in dedicated T2I models, yet still produces non-degenerate outputs across all six task categories at a level consistent with this parameter scale and training budget. This checkpoint serves as the T2I-initialized starting point for the I2L ablation in Table~\ref{tab:abl_t2i}.}
\label{tab:supp_t2i_geneval}
\begin{tabular}{l|cccccc|c}
& single & two & counting & colors & color & position & \textbf{overall} \\
& obj. & obj. & & & attr. & & \\
\shline
PixelART-B (BLIP-3o, T2I)
& 74.38 & 24.24 & 18.12 & 61.70 & 12.75 & 7.50 & \textbf{33.12} \\
\end{tabular}
\end{table}
 
For completeness, the prompt-level statistics are: total images 2{,}212, total prompts 553, correct images 31.87\%, correct prompts 53.16\%. Despite the T2I pretraining, the I2L fine-tuning loss curve in Figure~\ref{fig:abl_visualizations}(c) closes within $\sim$60K steps and the final I2L metrics in Table~\ref{tab:abl_t2i} are no better than those of the from-scratch baseline, supporting our finding that I2L can be learned end-to-end without external T2I priors at this scale.

\section{Generalization to Natural Images}
\label{app:natural_image}
 
\begin{figure}[!ht]
\centering
\includegraphics[width=\linewidth]{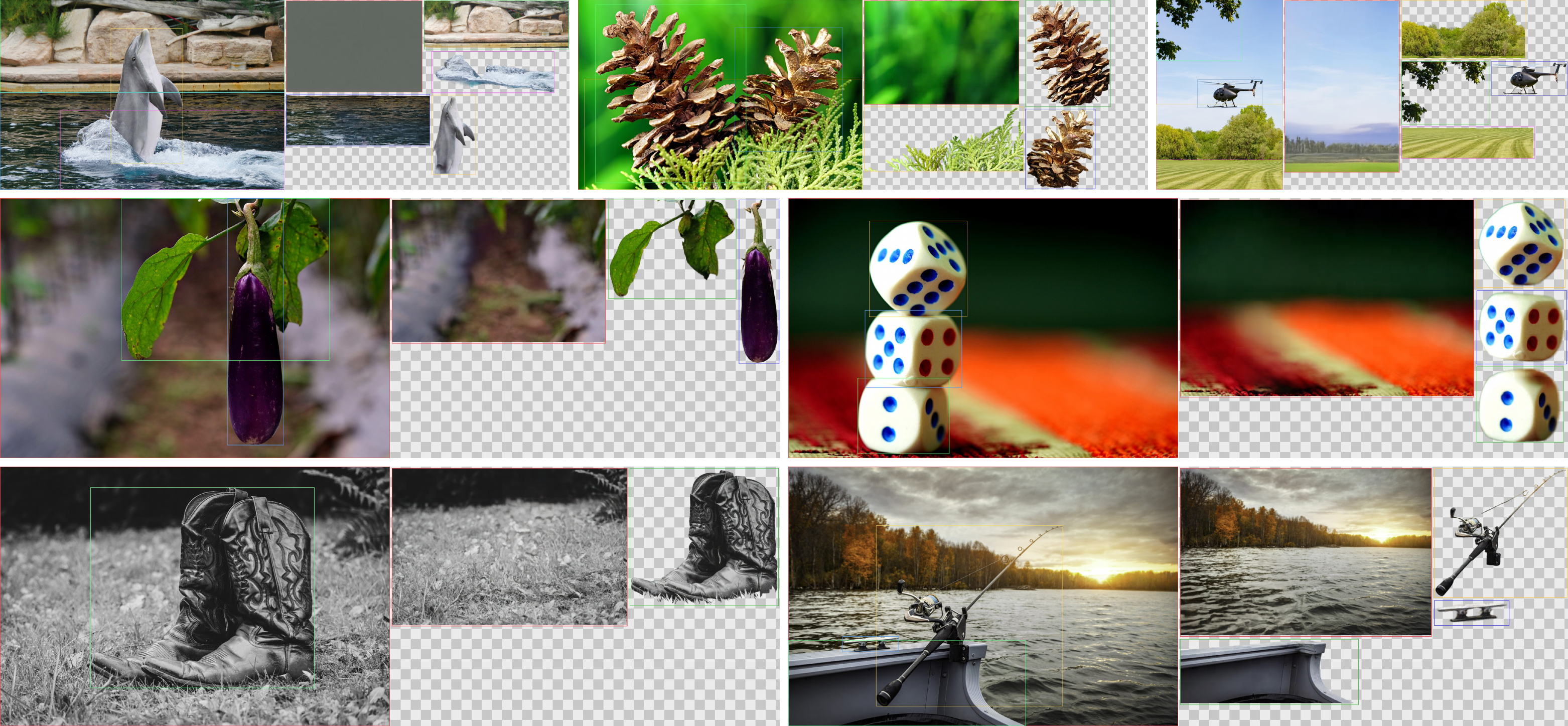}
\caption{\textbf{Zero-shot decomposition on natural images.} Although trained only on multi-layer design templates, PixelART produces coherent foreground/background separation and plausible occluded-region completion on selected natural-image inputs.}
\label{fig:natural_image_examples}
\end{figure}

PixelART is trained exclusively on multi-layer design templates. Although natural photographs lie outside this training distribution, the main model still produces meaningful decompositions on many natural-image inputs, suggesting that the I2L objective and pixel-space architecture together transfer beyond the design domain to a useful degree. Figure~\ref{fig:natural_image_examples} shows representative successful cases.
 
\section{Limitations}
\label{app:failure}
 
We summarize the most prominent limitations of PixelART in this section.
 
\begin{itemize}[leftmargin=2em, itemsep=2pt, topsep=2pt]
\item \textbf{Complex occlusion completion.} Filling in regions hidden behind large foregrounds is the model's clearest weakness, particularly on natural photographs. We attribute this to a coverage gap in our training distribution---multi-layer design templates rarely contain dense, photographic occlusions of the kind seen in natural scenes---rather than to a fundamental limit of the from-scratch pixel-diffusion architecture.
\item \textbf{Reduced reliability on natural images.} As shown in Appendix~\ref{app:natural_image}, the model can produce coherent decompositions on natural photographs, but the success rate is noticeably lower and less stable than on design content. Beyond the completion artefacts above, failures on natural inputs frequently take the form of \emph{layer assignment errors}: a near-empty foreground layer, residual foreground content in the background, multiple elements merged into one layer, or a single object split across layers. Closing this gap likely requires natural-image data with layered annotations during training.
\item \textbf{Unusually large $K$.} PixelART supports variable $K$ through anonymous regional tokens, but only up to a maximum of $K{=}32$ layers seen during training. For inputs that exceed or approach this limit, reconstruction precision drops noticeably and assignment errors become more frequent.
\item \textbf{Requires external layout.} PixelART consumes a bounding-box layout, supplied either by the user or by an automatic detector (Appendix~\ref{app:layout_detector}); it does not operate end-to-end on the input image alone, unlike bbox-free baselines such as Qwen-Image-Layered.
\item \textbf{Limited reproducibility from public assets.} Because the 4M training corpus and the internal layout detector are not publicly released, the full system cannot be exactly reproduced from public assets alone. To support independent verification, we will release evaluation scripts, model outputs on the public LICA benchmark, the corresponding VLM-judge grids and per-sample metrics, and the detector-predicted bounding boxes used in our quantitative tables. The full architecture and training recipe are specified in Appendix~\ref{app:config}, enabling reimplementation on alternative layered datasets.
\end{itemize}

\section{Broader Impacts}
\label{app:broader_impacts}
 
\paragraph{Positive impacts.} PixelART targets design and editing workflows. By recovering editable RGBA layer stacks from rasterized images, it reduces the manual cost of repurposing, retargeting, or updating visual content, and lowers the barrier to image editing for users without professional design experience.
 
\paragraph{Potential misuse.} As with other generative editing systems, PixelART can be misused. The most direct risks are: (i) image manipulation with deceptive intent, such as removing or altering disclaimers, context-providing details, or other visible information; (ii) decomposition of copyrighted design assets to facilitate unauthorized reuse; and (iii) removal of watermarks, signatures, or other provenance markers. These risks are bounded relative to full text-to-image generation, since PixelART operates on an existing input rather than fabricating new scenes from scratch, but they are direct enough to warrant explicit acknowledgement.
 
\paragraph{Mitigations.} For any production deployment of this technology, we recommend a small set of pre-input safeguards: (i) automatic detection of copyrighted or otherwise protected material in the input image, with processing refused on positive matches; and (ii) automatic rejection of requests whose bounding-box layout targets watermarks, signatures, or other provenance markers. We also encourage the broader community to develop forensic methods that can detect layer-level re-editing, complementing existing image-tampering detection work.

\section{External Assets and Licenses}
\label{app:assets}
 
We list the external assets used in this work, citing the original creators and the licenses under which the assets are released. All assets are used in accordance with their respective licenses.
 
\paragraph{Models.}
\begin{itemize}[leftmargin=2em, itemsep=2pt, topsep=2pt]
\item \textbf{Qwen3-VL-2B-Instruct}~\cite{bai2025qwen3}, used as the base model for the internal layout detector. Apache License 2.0.
\item \textbf{Qwen-Image-Layered}~\cite{yin2025qwenimagelayered}, used as a system-level baseline (Table~\ref{tab:compare_main}) and as the source of the RGBA-VAE encoder for the latent-space ablation (App.~\ref{app:vae_decoder}). Apache License 2.0.
\item \textbf{CLD}~\cite{cld}, used as a system-level baseline (Table~\ref{tab:compare_main}). License inherited from FLUX.1-dev: FLUX.1~[dev] Non-Commercial License.
\item \textbf{LayerD}~\cite{suzuki2025layerd}, used as a non-diffusion paradigm reference (Table~\ref{tab:compare_main}). Apache License 2.0.
\end{itemize}
 
\paragraph{Datasets and benchmarks.}
\begin{itemize}[leftmargin=2em, itemsep=2pt, topsep=2pt]
\item \textbf{LICA}~\cite{hirsch2026licalayeredimagecomposition}, used as an evaluation benchmark in Table~\ref{tab:compare_main}. CC BY 4.0.
\item \textbf{BLIP-3o pretraining corpus}~\cite{chen2025blip3o}, used for the T2I-pretraining ablation (App.~\ref{app:t2i_pretrain}). Apache License 2.0.
\item \textbf{GenEval}~\cite{ghosh2023geneval}, used to evaluate the T2I-pretrained checkpoint (App.~\ref{app:t2i_pretrain}). MIT License.
\end{itemize}
 
\paragraph{Closed external APIs.}
\begin{itemize}[leftmargin=2em, itemsep=2pt, topsep=2pt]
\item \textbf{GPT-5.5} (OpenAI), used as a VLM judge for layer quality scoring (App.~\ref{app:vlm_prompt}). Use is subject to OpenAI's API usage policies.
\item \textbf{Gemini-3.1 Pro} (Google), used as a second VLM judge. Use is subject to Google's Generative AI API terms of service.
\end{itemize}

\end{document}